\documentclass[12pt,onecolumn]{IEEEtranTCOM}

\normalsize

\usepackage[pdftex]{graphicx}
\usepackage{epstopdf}
\DeclareGraphicsExtensions{.pdf,.jpeg,.png,.eps}

\usepackage{color,balance}
\usepackage[cmex10]{amsmath}
\usepackage{amssymb}
\usepackage{array}
\usepackage{mdwmath}
\usepackage{amsthm}
\usepackage{makecell}
\usepackage{multirow}
\usepackage{booktabs}
\usepackage{comment}
\usepackage{graphicx}
\usepackage{textgreek}
\usepackage{float}
\usepackage{cite}
\usepackage{fixltx2e}
\usepackage{amsmath}
\usepackage{bm}
\usepackage{algorithm}
\usepackage{algpseudocode}
\usepackage{mdwtab}
\usepackage{tabularx}
\usepackage{xcolor}
\usepackage{xpatch}
\usepackage{adjustbox}

\usepackage{balance}
\usepackage{cite}

\allowdisplaybreaks

\newtheoremstyle{mystyle}
  {3pt}
  {3pt}
  {\itshape}
  {\parindent}
  {\bfseries}
  {\upshape{:}}
  {.5em}
  {}

\usepackage{color,soul}
\usepackage{setspace}

\theoremstyle{mystyle}

\usepackage[british,UKenglish]{babel}

\usepackage{subcaption}
\usepackage[font={small}]{caption}

\graphicspath{{./figures/}}

\usepackage{tabularx}
\usepackage{booktabs}
\usepackage{array}

\usepackage{subcaption}

\begin{document}

\title{RIS-Assisted Proactive Handover for Reliable mmWave Wireless Networks}

\author{Alaa Adnan,~\IEEEmembership{Graduate Student Member,~IEEE,} Mohammad Al-Quraan, ~\IEEEmembership{Graduate Student Member,~IEEE,} Ahmed Zoha,~\IEEEmembership{Senior Member,~IEEE,} M. Majid Butt, ~\IEEEmembership{Senior Member,~IEEE,}  Sami Muhaidat,~\IEEEmembership{Senior Member,~IEEE,} Muhammad Ali Imran,~\IEEEmembership{Fellow,~IEEE,} Marco Di Renzo,~\IEEEmembership{Fellow,~IEEE,} and Lina Mohjazi,~\IEEEmembership{Senior Member,~IEEE}
\thanks{A. Adnan, M. Al-Quraan, A. Zoha, M. A. Imran and L. Mohjazi,  are with the James Watt School of Engineering, University of Glasgow, Glasgow, G12 8QQ, UK. (e-mails: \ A.Bibi.1@research.gla.ac.uk, \{Mohammad.Alquraan, Ahmed.Zoha, Muhammad.Imran, Lina.Mohjazi\}@glasgow.ac.uk).}

\thanks{M. Majid Butt is with with Nokia Standards, USA. (e-mail: Majid.Butt@nokia.com).}

\thanks{S. Muhaidat is with the KU 6G Research Center, Department of Computer
and Communication Engineering, Khalifa University, Abu Dhabi 127788,
UAE, and also with the Department of Systems and Computer Engineering, Carleton University, Ottawa, ON K1S 5B6, Canada. (e-mail: sami.muhaidat@ku.ac.ae).}
\thanks{ M. Di Renzo is with Universit\'e Paris-Saclay, CNRS, CentraleSup\'elec, Laboratoire des Signaux et Syst\`emes, 3 Rue Joliot-Curie, 91192 Gif-sur-Yvette, France (e-mail:marco.di-renzo@universite-paris-saclay.fr), and with King's College London, Centre for Telecommunications Research Department of Engineering, WC2R 2LS London, UK. (e-mail:marco.di\_renzo@kcl.ac.uk).}

}
\maketitle
\markboth{}{}
\begin{abstract}
Millimeter-wave (mmWave) networks are highly susceptible to line-of-sight (LoS) blockages. Vision-aided wireless communications (VAWC) enable proactive handovers (PHO) to mitigate such blockages; however, PHO becomes challenging when no nearby base station (BS) is available. In such cases, reconfigurable intelligent surfaces (RIS) can be used to restore connectivity.
To ensure timely PHO, the RIS configuration time must be taken into account, as the large number of RIS elements can limit responsiveness in time-sensitive scenarios. This work proposes a novel RIS-assisted PHO approach that optimizes the number of allocated RIS elements to balance signal processing complexity and link quality under handover timing constraints, making the RIS-assisted link more energy-efficient. An optimization problem based on particle swarm optimization (PSO) is formulated to determine the optimal end-to-end RIS link setup that runs offline to bypass latency constraints.
Results show that reducing the number of RIS elements by 12\% leads to a 10\% decrease in dissipated energy without compromising the signal-to-noise ratio (SNR). Moreover, the RIS-assisted link achieves a 15--30~dB improvement in blocked regions while maintaining accurate PHO timing.

\end{abstract}
\section{INTRODUCTION} \label{intro}
 Millimeter-wave (mmWave) technology has emerged as a promising candidate for sixth-generation (6G) networks due to its ability to support higher transmission rates \cite{[2]}. Beamforming is envisioned to play a key role in mmWave systems, providing spatial reuse and reducing interference \cite{[2-1]}. However, in terms of reliability, beamforming-based mmWave connectivity mainly relies on directive line-of-sight (LoS) links due to channel sparsity, which limits the usefulness of multipath components \cite{[2-2]}. This makes mmWave signals highly susceptible to blockages, causing severe signal degradation when a LoS beam is obstructed.

Vision-aided wireless communication (VAWC) frameworks are expected to play a crucial role in 6G networks by leveraging visual information of the wireless environment to enable key applications including optimal beam selection based on user location \cite{[1]}, throughput estimation for choosing the best base station (BS) to serve a user \cite{[2-3]}, and, most importantly, beam blockage prediction for proactive handover (PHO) in mmWave networks \cite{[7]}. In many cases, a PHO to another BS may not always be possible due to availability or deployment constraints. Meanwhile, reconfigurable intelligent surfaces (RIS) have emerged as a promising technology and an affordable alternative to BS availability for enhancing signal quality and mitigating LoS blockages, as the RIS can steer the signal towards a desired location using RIS-assisted links \cite{[1-2]}.

In our previous work \cite{[3-3]}, we integrated these two concepts into a unified framework to enhance reliability and address the blockage issue. The findings revealed that to provide acceptable signal strength, the RIS-assisted link must utilize a large RIS array to improve the signal quality. The need for larger RISs is attributed to the nearly passive nature of the RIS and the fact that the effective aperture area of an RIS is typically smaller than its actual physical area, further increasing the demand for a larger RIS to enhance link performance. The work in \cite{[3-6]} introduced a mathematical model that accounts for the aperture area efficiency and proposed a modified path loss model based on the effective RIS area. Notably, the model considered the far-field assumption and estimated the RIS diameter to be approximately 20mm for a frequency of 28GHz, with an even smaller effective RIS area available for establishing a link. As a result, at higher frequency bands, the RIS effective area becomes even smaller. To overcome this limitation, achieving better performance requires deploying physically larger RISs. 

The need for physically larger RISs operating in the mmWave band necessitates denser RIS configurations composed of smaller elements, due to the shorter wavelengths at these frequencies. Since each RIS element typically has an area smaller or equal to half of a wavelength squared, this implies that a significant number of elements can be accommodated per unit area. For instance, the authors in \cite{[3-7]} fabricated and tested an RIS in both near-field and far-field scenarios. The fabricated RIS had an area of $10 \text{ cm} \times 10 \text{ cm}$, operated at 28GHz, and contained 400 RIS elements within this relatively small space. Deploying dense RISs operating in the mmWave band introduces several challenges, such extensive channel estimation, and complex beam training (BT), limiting the RIS application in real-time or time-sensitive applications. 

With large number of RIS elements, satisfying far-field conditions often requires distances that exceed the Fraunhofer distance. This condition may result in a scenario where the link between the BS and the RIS operates in the far-field region, while the link between the RIS and the user remains in the near-field region \cite{[3-4]}. The increased number of RIS elements in the near-field requires higher-dimensional beamforming vectors to account for the spherical wavefront effect of each RIS element, further increasing the complexity of channel estimation and BT.
Another important factor influenced by the high number of mmWave RIS elements is the growing demand for green technologies, a core requirement of 6G networks aiming for energy-efficient operation. Describing the RIS as a passive system is therefore not entirely accurate, as its configuration incurs power dissipation costs. Specifically, this includes the static power dissipation required to operate the RIS controller layer and the surface power dissipation associated with driving the diodes embedded in each RIS element used for beam steering~\cite{[18]}. Efficiently minimizing the number of RIS elements required to establish a reliable link can significantly reduce overall power consumption and enable more sustainable RIS deployments. For instance, self-sustainable RISs \cite{[19]} are envisioned for installation on structures such as building façades, where energy sources may be limited or difficult to access. In such scenarios, utilizing a smaller number of RIS elements becomes essential to align RIS operation with green communication objectives and the sustainability goals of 6G networks.

\subsection{Related Work} \label{related}

In this section, we present state-of-the-art research that proposes solutions to beam blockages in high-frequency bands.

\subsubsection{Beam Blockage Prediction}

Sensing based beam blockage prediction plays a crucial role in ensuring reliable communication, as it triggers the handover process to a link with better conditions than the blocked one various data acquisition techniques are employed for link blockage prediction, including VAWC frameworks, light detection and ranging (LiDAR) sensing, and radar-based blockage prediction. For instance, the work in \cite{[5]} leveraged a VAWC framework, using a set of RGB images with predefined beamforming vectors to train a deep learning (DL) model for blockage prediction in mmWave networks. The work in \cite{[8]} proposed the use of RGB images with depth cameras to train a machine learning (ML) model called adaptive regulation of wave vectors (AROW), facilitating proactive BS selection in blockage scenarios. The selection criterion was based on throughput estimations to ensure the best link for a handover. Additionally, \cite{[R-3]} utilized LiDAR to acquire point cloud data for mmWave link quality prediction, forecasting radio propagation fluctuations caused by pedestrian obstructions. The work in \cite{[R-4]} demonstrated that integrating radar-based sensing with mmWave BSs provides valuable information, such as velocity and range, which can aid in predicting network obstacles.

\subsubsection{ Beam Blockage Mitigation by RIS-Assisted Links}

RIS-assisted links provide an effective solution for replacing blocked LoS links by redirecting incident signals to obstructed users. The potential to reduce channel estimation and beam training overhead has garnered significant interest in enhancing the practical deployment of RISs, particularly for time-critical or real-time applications.  

The study in \cite{[R-5]} aimed to optimize RIS performance by jointly designing BS beamforming and RIS reflection coefficients. The proposed approach maximized the user's sum rate by directly optimizing received pilots at the BS using a graphical neural network (GNN) learning model. This method eliminated the need for explicit channel estimation. Results demonstrated that incorporating additional information, such as user location reduced pilot transmission achieving optimal beamforming and RIS reflection settings. Despite these improvements, the complexity of the beamforming matrix design remains directly proportional to the number of RIS elements, leading to increased training delays. This led the authors to assume that GNN training is performed offline to mitigate the impact of runtime delays in online scenarios.  
Another promising approach involves intelligent RIS beam management (BM) and BT instead of explicitly estimating channel state information (CSI). For example, \cite{[R-7]} trained a ML model within a BM framework to process environmental and user mobility data, enhancing BM in beam blockage scenarios. The work in \cite{[R-6]} employed BT to estimate the best beam selection for the receiver. However, this method heavily depended on RIS phase shifter resolution, often assuming an ideal continuous phase shifter capable of selecting the optimal beam with minimal directivity loss.
Furthermore, \cite{[13]} proposed a joint active and passive beamforming design for optimal beam selection in RIS-assisted networks, utilizing the concept of a channel knowledge map (CKM). This approach builds a site database that correlates beams with users' geographical locations, aiding in optimized beam selection. Also, this work highlighted the fact that BT overhead is directly related to the number of RIS elements, and with the large number of mmWave RIS elements, BT will be a challenging task.
\subsubsection{RIS-Assisted HO}

In our work~\cite{[R-8]}, we presented and discussed the challenges faced by RIS-assisted handover (HO) in mmWave networks, particularly the limitations of traditional reactive HO approaches, where the HO procedure is triggered based on reported measurements from the user equipment (UE). Such mechanisms allow link degradation to occur before initiating the HO process. However, this approach does not satisfy the stringent requirements of 6G ultra-reliable low-latency communication (URLLC) services, such as autonomous vehicle driving and telemedicine, where seamless and reliable connectivity is essential. Therefore, proactive approaches represent a more suitable solution for such scenarios.

The work in~\cite{[R9]} proposed a blockage-aware RIS-assisted HO framework based on DL model, where the model learns channel and blockage conditions from previously observed patterns to reduce unnecessary HO events. Similarly, the work in~\cite{[R10]} introduced a Neyman-Pearson (NP)-based blockage detection mechanism to enable robust RIS beamforming schemes for minimizing outage probability. However, most existing RIS-assisted blockage-aware frameworks are primarily predictive rather than truly proactive, since they mainly depend on learned historical patterns or communication-layer measurements. In contrast, true PHO frameworks should rely on real-time processed sensing data to proactively maintain seamless connectivity and satisfy URLLC requirements.

Despite recent advancements in blockage prediction and mitigation techniques, the integration of RIS with VAWC framework for seamless RIS-assisted handover has not been sufficiently explored in prior studies, as most existing works assume the availability of an alternative BS to which the blocked user can be handed over. However, in practical deployment scenarios, a secondary BS may not always be available. Additionally, most RIS-related studies assume that user locations and blockage information are either perfectly known at the BS or modeled probabilistically, which may not be suitable for highly dynamic real-time environments. Another overlooked factor is the impact of RIS configuration time on system performance and handover preparation delay.

Moreover, even BM and beam BT techniques, mmWave RIS implementations still require a large number of RIS elements to achieve acceptable communication performance. In our previous study~\cite{[3-3]}, we performed a system-level performance analysis for a mmWave RIS-assisted link operating at 60~GHz. The results indicated that at least 1000 RIS elements are required to establish a communication link with sufficient signal strength. Consequently, signal processing complexity and beam training overhead significantly increase with RIS size. Therefore, optimizing the RIS-assisted handover process by minimizing the number of allocated RIS elements while maintaining acceptable link quality is crucial for practical deployment.

\subsection{Motivation and contribution}
This work aims to leverage an RIS-assisted link within a PHO framework to replace a blocked LoS connection. However, deploying large-scale RIS-assisted mmWave communication systems introduces several challenges, which become critical bottlenecks for time-sensitive applications such as PHO. In particular, RIS configuration time must be explicitly accounted for as part of the handover preparation phase among the three standard 3GPP handover stages, namely preparation, execution, and completion as proposed in our work \cite{[R-8]}, since proactive handover operation is constrained by narrow time windows for triggering and completing the handover process before severe link degradation occurs.  

To effectively leverage RIS for PHO, two key deployment challenges must be addressed:

\begin{itemize}  
    \item Ensure that the RIS has a sufficient number of elements to maintain the required link quality.  
    \item Ensure that the time needed to configure the phase shifts of the RIS elements is within the acceptable time limit to trigger handover before the occurrence of a blocking event.  
\end{itemize}  
From a practical deployment perspective, finding a trade-off between achieving the desired RIS-assisted link quality and mitigating signal processing complexity is crucial. This trade-off can be realized by efficiently minimizing the number of allocated RIS elements while maintaining acceptable link quality. Such an approach not only reduces the complexity of channel estimation, energy dissipation, and BT overhead but also enhances the feasibility of RIS-assisted links for real-time applications. Since PHO depends on accurate timing to trigger the process, the time required for RIS configuration becomes a critical factor in ensuring a successful PHO. However, no studies have specifically focused on the impact of RIS configuration time on PHO timing, especially during the HO preparation phase. In the surveyed literature, RIS configuration time was either estimated or assumed for purposes unrelated to PHO considerations.  

The work presented in \cite{[3-15]} suggested that RIS configuration response times need to fall within a granular range of 20 to 100 milliseconds (ms) to support effective communications. The study in \cite{[3-16]} employed the STM32L071V8T6 microcontroller unit to control RIS phase shift configurations, reporting a configuration time of less than 35~ms. Meanwhile, the work in \cite{[3-7]} utilized a field-programmable gate array (FPGA) for RIS phase shift control, where the configuration delay ranged from 0.22~ms to 7~ms, depending on the number of RIS elements. These measurements also used to indicate wiring complexity in RIS fabrication, highlighting the challenges of managing more mmWave RIS elements, including greater fabrication difficulty, control complexity, and longer configuration time.
In these studies, the RIS configuration time was not considered within the context of a communication framework such as PHO. A deeper understanding of the factors influencing RIS configuration time is essential for the effective and practical deployment of RIS-assisted links in time-critical operations such as handovers and user tracking. To the best of the authors' knowledge, no prior work has addressed the impact of RIS configuration time on PHO in RIS-assisted links.

mmWave RIS-assisted systems are envisioned to play a significant role in 6G networks, enabling the realization of smart environments (SE) supporting applications such as road safety and autonomous vehicles in ultra dense networks (UDNs) all of which demand seamless and uninterrupted connectivity. However, all the challenges mentioned earlier are directly dependent on the number of RIS elements used to establish a link, thereby hindering the practical implementation of RIS-assisted systems. Addressing the task of achieving the balance between link quality and reduced signal processing is essential to enable efficient and practical RIS-assisted link deployment capable of replacing blocked mmWave LoS connections within a PHO framework. In this context, the key contributions of this work are summarized as follows:

\begin{itemize}
    \item Integration of RIS-assisted mmWave connectivity within a novel VAWC framework enables blockage prediction and precise user localization, thereby facilitating accurate beam steering from the RIS toward users in blocked regions using practical RIS beamforming with quantized phase shifts.

    \item Development of an optimization algorithm to jointly enhance the responsiveness of the end-to-end RIS-assisted PHO process by minimizing the number of allocated RIS elements to reduce configuration time and RIS dissipated energy while maintaining the desired link quality. The optimization process investigates the optimal combination of parameters such as the transmitter MIMO array requirements, the best set of beams for the cascaded channel, and the allocated number of subcarriers. 
    
    \item Proposal of a novel RIS-assisted PHO framework that explicitly incorporates RIS configuration time within the handover preparation phase of the standard 3GPP handover procedure, enabling seamless and timely handover triggering during LoS blockage scenarios.

\end{itemize}

The rest of this paper is organized as follows: In Section \ref{sys}, the system and channel models are presented. Section \ref{prob} discusses the problem formulation and optimization approach. Section \ref{VAWC} introduces the  proposed RIS-assited PHO framework. Section \ref{results} provides the results and key findings. Finally, Section \ref{conc} gives concluding remarks.

\section{System and channel models}\label{sys}  
In this section, we provide a comprehensive description of the RIS architecture and configuration time estimation, which will be integrated into the VAWC PHO framework, along with the system and channel models.
 
\subsection{RIS Configuration Time Estimation} \label{RIS}
\begin{figure}
\centering
  \includegraphics[scale=0.35]{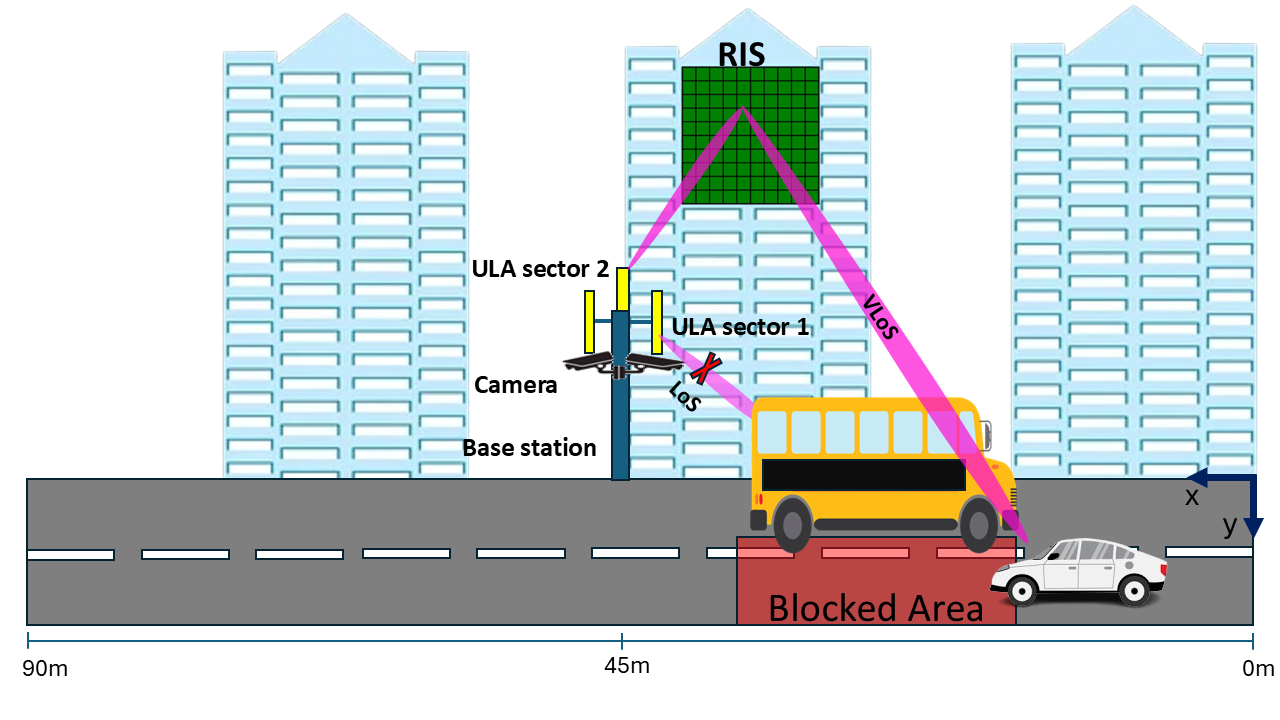} 
 \caption{RIS-assisted vision-aided BS setup.}
 \label{fig:4}
\end{figure}

In order to effectively deploy an RIS-assisted link in a PHO procedure, the RIS configuration time required to steer the signal toward a blocked area must be accounted for to ensure accurate triggering and seamless connectivity. 
This work builds upon our previous study \cite{[3-3]}, which proposed a PHO framework that leverages RIS-assisted beamforming in a blockage scenario (see Fig.~\ref{fig:4}). The RIS-assisted beam is selected from a pre-stored set of RIS beamforming configurations maintained in a CKM database. However, the PHO framework in \cite{[3-3]} did not account for RIS configuration time as part of the PHO timing scheme, which is a critical factor for ensuring accurate PHO execution.
The framework proposed in \cite{[3-3]} was based on the original PHO framework introduced in \cite{[7]}, which utilized VAWC and RGB cameras powered by object detection algorithm to detect blockages. The detected blockage data was then used to train a neural network (NN), which relied on user mobility speed to determine the optimal triggering point for the PHO process. However, the framework in \cite{[7]} did not consider PHOs involving an RIS. Instead, it proposed performing a PHO to a secondary BS when a blockage occurs. In practice, a secondary BS may not always be available as explained in section \ref{intro}, making an RIS-assisted link a promising and practical alternative for handover and, thus, worth further investigation.

In the original framework proposed by \cite{[7]}, two important timing factors were estimated: the execution time \(T_{\text{exec}}\) and the waiting time \(T_{\text{W}}\) before triggering the handover. \(T_{\text{exec}}\) is defined as the time required for the proposed algorithm to be completed, starting when the images are captured by the BS and processed for blockage detection until the PHO process is completed. The value of \(T_{\text{W}}\) is given as follows \cite{[7]}

\begin{equation}
T_{\text{W}} = T_{\text{to-BLK}} - T_{\text{exec}} \; ,
\label{eq:3}
\end{equation}
where \( T_{\text{to-BLK}} \) represents the time for a user to encounter a blockage. \( T_W \) varies based on the user’s location and movement speed. Selecting the value of \( T_W \) is critical for defining the optimal PHO trigger region, allowing the algorithm to determine the best time to perform the PHO.
In order to integrate the RIS as part of the PHO framework \( T_{\text{exec}} \) must include the RIS response time \( T_{\text{c}} \) representing the required RIS configuration time to steer the signal to a blocked user. As a result, the updated execution time, denoted as \( T^{\text{RIS}}_{\text{exec}} \), can be expressed as
\begin{equation}
T^{\text{RIS}}_{\text{exec}} = T_{\text{exec}} + T_{\text{c}} \; ,
\label{eq:4}
\end{equation}

This leads to a new waiting time, \( T^{\text{RIS}}_{\text{W}} \), which is given by

\begin{equation}
T^{\text{RIS}}_{\text{W}} = T_{\text{to-BLK}}- T^{\text{RIS}}_{\text{exec}} \; ,
\label{eq:5}
\end{equation}

The new \( T^{\text{RIS}}_{\text{W}} \) now includes the RIS configuration time \( T_{\text{c}} \), in order to be able to estimate \( T_{\text{c}} \) we examine the RIS architecture and identify the factors contributing to the configuration time.

The design approach for RIS can vary based on multiple factors, such as indoor or outdoor deployment, operating frequency, RIS element design, and several other considerations. This work does not focus on the specific details of the radio frequency design aspects of RIS. Instead, it emphasizes the control layer setup and its impact on the configuration time. The RIS consists of three main layers \cite{[3-7]} the front layer, which is a grid of RIS elements; the RIS back-layer board; and the control layer. The front layer of the RIS board typically consists of \(N\) RIS elements, each embedded with \(B\) diodes to control the phase shift of the incident signal for beam steering. The most commonly deployed diodes include varactor diodes \cite{[3-17]} and PIN diodes \cite{[3-18]}.
The control layer is responsible for transmitting configuration bits to the RIS back-layer board. The back layer of the RIS board houses a set of logic circuits that control the diodes embedded in each RIS element, enabling them to switch on or off as needed \cite{[R8]}. This layer typically contains multiple shift registers, where the number of registers is generally \( \leq N \), depending on the number of diodes embedded in each element and the shift register bus capacity. Since each diode requires one bit to be switched on or off, the shift registers are used by the controller layer to push the configuration bits via serial-to-parallel transmission, efficiently controlling the diode states. Different setups of shift registers controlling the RIS elements are presented in \cite{[3-7]}, \cite{[3-18]}, and \cite{[3-19]}.

The configuration time required for the RIS depends on several factors, including control signaling, memory access, hardware coordination, the transmission rate of the control board, the clocking rate of the shift registers, the number of RIS elements, and the number of diodes embedded within each RIS element~\cite{[3-7]}. Thus, the codeword, which represents the total number of bits required to configure the RIS, is determined by multiplying the total number of RIS elements by the number of diodes contained in each element. Each bit is used to switch a single diode on or off, thereby enabling the desired RIS beamforming configuration.

When configuring the RIS remotely from a server, the interface transmission rate directly impacts the overall configuration time. However, if the configuration states are stored locally within the microcontroller layer and the required codeword is directly pushed to the shift registers, the configuration time depends primarily on the clocking rate of the shift registers~\cite{[3-7]}. The work presented in this paper adopts the latter approach. Therefore, the remaining delays, including control signaling, memory access, and hardware coordination, are considered fixed overheads that can be measured and incorporated into the overall system timing during practical deployment. Accordingly, the required RIS configuration time can be expressed as
\begin{equation}
T_{\text{c}} = \frac{N B}{C_{lk}}   \
 ,
\label{eq:1}
\end{equation}
where \( \ C_{lk} \) represents the clocking rate of the shift registers and in our work it is assumed to be \( 100 \, \text{kHz} \)~\cite{[3-7]}.
From Eq. (\ref{eq:1}), together with the additional delays accounted for in Eq. (\ref{eq:5}), it follows that achieving a faster RIS configuration response requires minimizing \( N \). 
Another important aspect that is governed by \(N\) and \(B\) is the dissipated power required to drive the diodes for the RIS to achieve the required configuration. Each time a diode changes state from 0 to 1 or vice versa, the dissipated power \(P_{\mathrm{d}}\) is given by \cite{[24]}   
\begin{equation}
P_{\mathrm{d}} = N B C_{\mathrm{l}} V_{\mathrm{cc}}^{2} C_{\mathrm{lk}}.
\label{pd}
\end{equation}
Here, \(C_{\mathrm{l}}\) is the load capacitance of each diode, and \(V_{\mathrm{cc}}\) is the supply voltage used to operate the shift registers.
By identifying the optimal value of \( N \), we can estimate the RIS configuration time. Efficiently minimizing \( N \) without compromising link quality can reduce channel estimation, BM and BT complexity, as well as lower the dissipated power, since all of these factors depend on \( N \). As described in equations (\ref{eq:1}) and (\ref{pd}), \( N \) is a key parameter in estimating \( T_{\text{c}} \), given that both \( B \) and \( C_{lk} \) are fixed for each element and the shift register. The optimization approach will be further explored in Section \ref{prob}, where the proposed algorithm to achieve this goal is introduced.

\subsection{System Model}
We consider a BS operating in the mmWave band, assisted by an RIS, where the BS transmits to a single-antenna user. The BS coverage area is served by multiple uniform planar array (UPA) sectors, with each UPA responsible for a specific portion of the service region. Each UPA sector consists of \( M \) antenna elements and employs hybrid beamforming, utilizing one or more subarrays \( m_{\text{sub}} \) to transmit to the user. The total number of radio frequency (RF) chains is given by \( N_{\text{rf}} = M / m_{\text{sub}} \) \cite{[20]}. 
By feeding the same baseband signal to multiple RF chains, the system can achieve higher array gain depending on the connection conditions and performance requirements, the group of dedicated RF chains results in an array expressed as \( M_{\text{active}} = N_{\text{active}} \times m_{\text{sub}} \), where \( N_{\text{active}} \) is the number of RF chains used for transmission to the user and \( N_{\text{active}} \leq N_{\text{rf}} \). 

The system employs orthogonal frequency-division multiplexing (OFDM) with \(K\) subcarriers as a resource allocation scheme for radio access. In the analog beamforming stage, the mmWave BS system uses a UPA, where each beamforming vector is selected from a predefined codebook \( \mathbf{\mathcal{F}} = \{f_i\}_{i=1}^{B_m} \), with each vector constructed as \( f_i = f_{i,x} \otimes f_{i,z} \), where \(f_{i,x}\) and \(f_{i,z}\) are the horizontal and vertical steering vectors, respectively, following the approach in \cite{[20]}. Here, \(\otimes\) denotes the Kronecker product, which forms a 2D beamforming vector by taking all possible element-wise products between the horizontal and vertical steering vectors. \(B_m\) represents the total number of beams in the codebook.

To monitor the environment, the BS is equipped with a set of cameras that oversee the street, as illustrated in Fig.\ref{fig:4}. These cameras are used to determine the user’s location and detect potential blockages, which is essential for identifying the optimal beam to serve the user \cite{[3-13]}. This study adopts the blockage scenario from the original dataset \cite{[16]}, where co-located cameras detect a bus obstructing the LoS signal to the served user. We extend the system setup by introducing an RIS, as illustrated in Fig. \ref{fig:4}, to address anticipated beam blockages within the VAWC framework proposed in \cite{[7]}. Instead of performing a handover to a secondary BS, as suggested in \cite{[7]}, this work proposes establishing a virtual line-of-sight (VLoS) link using the RIS in case a second BS is not available. This VLoS link is leveraged by the PHO procedure as an alternative to the obstructed LoS link.
\begin{figure}
    \centering
    \includegraphics[width=0.3\columnwidth, keepaspectratio]{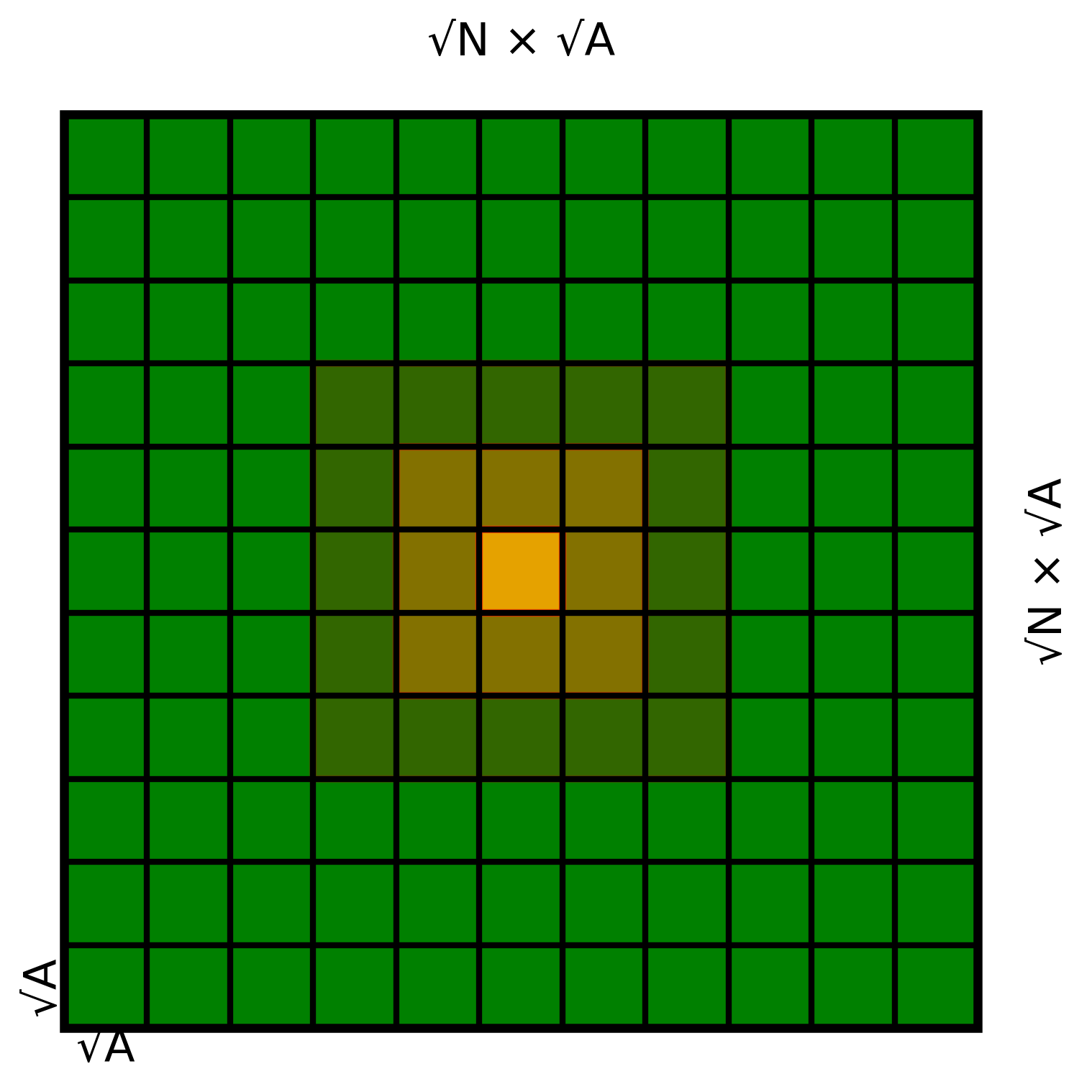}
    \caption{RIS structure with \( \sqrt{N} \) elements per side, full RIS area \( N \times A \), amber-shaded sub-squares represent centered subarrays with an odd number of elements per side, preserving a fixed RIS center across all allocation levels.}
    \label{fig:5}
\end{figure}
The RIS is comprised of elements, each of an area given by 
\( A = \left(\frac{\lambda}{2}\right)^2 \). The dimensions of each element are (\( \sqrt{A} \times \sqrt{A} \)), the elements are arranged in a planar square grid of size 
(\( \sqrt{N} \times \sqrt{A} \)). The total area of the RIS is (\(N \times A\)) \cite{[3-10]}. 

The RIS controller is responsible for element allocation to establish a VLoS link based on the required link demands. Several approaches for element allocation have been presented in the literature, such as in \cite{[21]}, where the proposed method involves allocating RIS elements starting from one corner in the form of a sub-square grid, progressively adding more elements outward until the entire RIS area is utilized. However, this approach may lead to several issues, as the center of the effective RIS array continuously shifts depending on the allocated elements. This shift results in a significant reduction in the aperture efficiency of the RIS, and each change in the number of allocated RIS elements may cause misalignment, requiring a different beam from the BS to the RIS to maintain maximum gain. 
In contrast, we envision an allocation approach based on square subarrays that start from the center of the RIS and expand outward, ensuring a unified RIS center across all allocation levels. To achieve this, the condition must be that the RIS side length \( \sqrt{N} \) is an odd number \cite{[3-4]}, and the allocated subarrays must also be squares with an odd number of RIS elements per side, this approach is illustrated in Fig. \ref{fig:5}.

We assume the RIS is mounted on the façade of a building located behind the BS at a sufficient height to overcome obstructions. The RIS can dynamically steer the beam using a predefined $N \times N$ codebook, denoted by \( \mathcal{V} \), comprised of steering coefficients expressed as
\begin{equation}
v_{mn} = \alpha e^{j\frac{2\pi d_{\scalebox{0.45}{RIS}}}{\lambda}\sin\theta_{mn}} ,\label{eq:7}
\end{equation}
where \( j = \sqrt{-1} \), \(v_{mn} \in \mathbb{C}^{N \times N}\), \(\alpha\) is the reflection coefficient of each RIS element, and \(m, n = 1, \dots, N\). Additionally, \(d_{\text{RIS}}\) represents the RIS element spacing, and \(\theta_{mn} \in [0, 2\pi]\) denotes the RIS steering angle. 

To facilitate the PHO, each UPA sector is equipped with a camera to capture visual information. This data is processed by the VAWC framework to extract critical parameters for predicting a blockage event in advance, such as the existence of a blocking object, the user's speed toward the blocked area, and the estimated time and location of the blockage. Based on this information, the VAWC framework enables the BS to steer the RIS beam toward the anticipated blockage area, leveraging the PHO framework in \cite{[3-3]} to define the optimal time for triggering the PHO.

\subsection{LoS Channel Model} 
The user transmits a pilot signal, which is used to train the beam set $B_m$ and select the optimal beam that maximizes the signal-to-noise ratio (SNR). This work adopts the pilot signal/beam training methodology described in \cite{[3]}. Once the optimal beam is identified, it is utilized for downlink communication. The baseband mmWave signal transmitted from the BS to the user over a LoS link, in the absence of a blockage, is expressed as
\begin{equation}
     y_{\text{los}, k} = \textbf{G}_{\text{los}, k} \textbf{f}_{\text{los}, k}^{\star} s_{\text{los}, k} + n_{\text{los}, k}   
 \label{eq:8}\hspace{3pt},    
\end{equation}
where, \(s_{\text{los}, k}\) represents a transmitted symbol, \(n_{\text{los}, k} \sim \mathcal{CN}(0, \sigma^2)\) represents the additive white gaussian noise (AWGN) for the \(k\)-th subcarrier with zero mean and variance of \( \sigma^2 \), \(\textbf{G}_{\text{los}, k}\) is the channel gain which is expressed as \cite{[7]}
\begin{equation}
\textbf{G}_{\text{los}, k} = \sqrt{\beta_{k}} \, (\textbf{h}_{k})^{T} e^{\left(j 2 \pi \frac{k d}{K\lambda}\right)} f(\theta, \phi)_{k}
\label{eq:9} \hspace{3pt},
\end{equation}

\noindent In Eq. (\ref{eq:9}), \(\beta_{k}\) represents the large-scale fading coefficient, $f(\theta, \phi)_{k}$ is the function of the angles of departure and arrival, respectively, where \(\theta\) is the azimuth angle and \(\phi\) is the elevation angle. \(h_{k} \in \mathbb{C}^{M \times K}\) is the small-scale fading channel coefficient, \(T\) denotes the transpose of a vector. It is worth mentioning that, in this work, we consider a single-path channel model, a well-established approach commonly adopted for mmWave systems. At high operating frequencies such as 60~GHz, the contribution of multipath components is very limited, particularly in mmWave RIS-assisted links, where highly directional beams and strong path loss for none-line-of-sight (NLoS) components make the LoS path dominant. This assumption is further supported by the inherent sparsity of mmWave channels, where the majority of the received power is typically concentrated in a single dominant path due to high directivity and severe attenuation of NLOS components. Consequently, all simulations in this work are conducted under a single-path scenario \cite{[2-2]}.

Since the objective of this work is to minimize the number of RIS elements, i.e., the overall RIS area, to reduce configuration time, the channel gains must account for both the transmitter and RIS areas as functions of the number of antenna elements. Thus, \(\beta_{k}\) is modeled as a function of the transmitter and receiver antenna areas, as described in \cite{[3-11]}

\begin{equation}
\beta_k = \frac{A_{\text{t}} A_{\text{u}}}{(\lambda d_{BU})^2} \label{eq:10} \hspace{3pt},
\end{equation}
where $A_{\text{t}}$ is the transmitter antenna area, $A_{\text{t}}$ can be expressed as \(A_{\text{t}} = M \times A_{\text{MIMO}}\), where \(A_{\text{MIMO}}\) denotes the area of a single antenna element. Similarly, \(A_{\text{u}}\) corresponds to the area of a single receiver antenna element. In this work both  $A_{\text{MIMO}}$ and $A_{\text{u}}$ are set to have an area of \( \left(\frac{\lambda}{2}\right)^2 \). Additionally, \(d_{BU}\) is the distance between the BS and the user. 

The optimal beam, selected to establish the best link to the user at a given location, is determined as follows \cite{[7]}
\begin{equation}
\textbf{f}_{\text{los}, k}^{\star} = \underset{\textbf{f} \in \textbf{F}}{\text{argmax}} \frac{1}{K} \sum_{k=1}^{K} \rho \left| \mathbf{\textbf{G}}_{\text{los}, k} \mathbf{f} \right|^2 \label{eq:11_snr}\hspace{3pt}.
\end{equation}

\noindent where \(\rho\) represents the power-to-noise ratio, defined as \(\frac{P_{\text{t}}}{\sigma^2}\) with $P_{\text{t}}$ being the transmitted power. Based on this equation, the SNR \(\gamma_{\text{los}}\) is evaluated as follows \cite{[16]}\cite{[22]}
\begin{equation}
\gamma_{\text{los}} = \frac{1}{K} \sum_{k=1}^{K} \rho \left| \mathbf{G}_{\text{los}, k} \mathbf{f}_{\text{los}, k}^{\star} \right|^2 \label{eq:12}\hspace{3pt}.
\end{equation}

\subsection{VLoS Channel Model} 
Whenever a blockage event is detected by the VAWC framework, an RIS-assisted link is utilized to perform a PHO \cite{[3-3]}. The signal received through the RIS beam can be expressed as
\begin{equation}
y_{\text{VLoS}} = {(\textbf{v}^{\star})}^H \textbf{G}_{\text{VLoS}} \textbf{f}_{\text{BI}}^{\star} s_{\text{VLoS}} + n_{\text{VLoS}} \label{eq:13}\hspace{3pt},
\end{equation}
\noindent where, \(s_{\text{VLoS}}\) reresents the transmitted symbol through the VLoS link,  \({(\textbf{v}^{\star})}^H \textbf{G}_{\text{VLoS}} \textbf{f}_{\text{BI}}^{\star}\) denotes the total cascaded end-to-end channel from the BS to the blocked user through the RIS, \(\textbf{f}_{\text{BI}}^{\star}\) represents the optimal beam from the BS to the RIS, and \(\textbf{v}^{\star}\) denotes the optimal RIS steering vector that maximizes \(SNR\). Additionally, \(H\) refers to the Hermitian transpose operation \cite{[13]} \cite{[13-1]}.

In order to capture a practical system performance we adopt the scenario mentioned earlier in section \ref{intro} where the link from the BS to RIS falls in the far-field model and the link from the RIS to the user falls in the near-field model thus \( \textbf{G}_{\text{VLoS}}\) is comprised of two channel gains. The first characterizes the link from the BS to the RIS and is expressed by the far-field model similar to the Eq. (\ref{eq:10}) as follows
\begin{equation}
\beta_{\text{BI}} = \frac{A_{\text{t}} A_{\text{RIS}}}{(\lambda d_{\text{BI}})^2} 
\label{eq:14}\hspace{3pt},
\end{equation}
\noindent \(A_{\text{RIS}}\) is  the RIS area and is expressed as \(A_{\text{RIS}} = N \times A\), where \(A\) represents the area of a single RIS element, and \(d_{\text{BI}}\) is the distance between the BS and the RIS \cite{[3-10]}. 
\noindent For the second part of the cascaded channel, specifically the link from the RIS to the blocked user, the channel is assumed to be in the near-field model and can be expressed as \cite{[3-11]}
 
\begin{equation}
\beta_{\text{IU}} = \frac{4 A_{\text{RIS}}}{3 \lambda^2} \left( \frac{\sqrt{2 \left( \frac{(d_{\text{IU}})^2}{A_{\text{u}}} \right)}}{\sqrt{1 + 2 \left( \frac{(d_{\text{IU}})^2}{A_{\text{u}}} \right)} \left( 1 + 4 \left( \frac{(d_{\text{IU}})^2}{A_{\text{u}}} \right) \right)} \right. \\
\quad + \left. 2 \cot^{-1} \left( \sqrt{8 \left( \frac{(d_{\text{IU}})^2}{A_{\text{u}}} \right) \left( 1 + 2\frac{(d_{\text{IU}})^2}{A_{\text{u}}} \right)} \right) \right) \label{eq:15}\hspace{3pt},
\end{equation}
\noindent here, \(d_{\text{IU}}\) represents the distance from the RIS center to the blocked user. as \(d_{\text{IU}}\) becomes large and exceeds the Fraunhofer region then \(\beta_{\text{IU}}\) can be obtained from Eq. (\ref{eq:14}) as well. The final channel coefficients of the cascaded channel \(G_{\text{VLoS}}\) can be expressed as
\begin{equation}
\textbf{G}_{\text{VLoS,k}} = \sqrt{\beta_{\text{IU}}\beta_{\text{BI}}}  \textbf{h}_{c,k}   e^{\left(j 2 \pi \frac{ k(d_{\text{RIS}} + d_{\text{MIMO}})}{K\lambda}\right)}  f(\theta, \phi)  \label{eq:16}\hspace{0pt},
\end{equation}

\noindent where \(\textbf{h}_{\text{c,k}}\) is the small-scale fading coefficient for the cascaded channel. The SNR of the far-field RIS VLoS link can be expressed as
\begin{equation}
\gamma_{\text{VLoS,far}} = \frac{1}{K} \sum_{k=1}^{K} \rho \cdot \left|(\textbf{v}^\star)^H \textbf{G}_{\text{VLoS},k}\textbf{f}_{\text{BI}}^{\star} \right|^2, \label{eq:17}
\end{equation}
The near-field SNR can be expressed as follows 
\begin{equation}
\gamma_{\mathrm{VLoS,near}} = \frac{1}{K}\sum_{k=1}^{K}\rho\;
\Big|
\sqrt{\beta_{\!IU}\,\beta_{\!BI}}\;
\mathbf{h}_{\mathrm{BI},k}^H \mathbf{f}_{\rm BI}^\star\,
e^{-j\frac{2\pi k  d_{\mathrm{MIMO}}}{K\lambda}}\\
\times
\sum_{n=1}^{N}
h_{{\rm IU},n,k}\,
e^{\,j\!\left(\psi_n - \frac{2\pi k r_n}{K\lambda}\right)}
\Big|^2,
\label{eq:17near}
\end{equation}
where \(h_{BI,k}\) , \(h_{IU,n,k}\) represent the small-scale fading for the BS to RIS and RIS to user respectively, \(\beta_{\text{IU}}\) obtained from Eq. (\ref{eq:15}),  and \(r_n\) is the distance from the \(n_{th}\) RIS element to the user expressed as 
\begin{equation}
\begin{split}
r_{\text{n}} = \sqrt{(x_{\text{n}} - x_{\rm U})^2 
+ (y_{\text{n}} - y_{\rm U})^2 
+ (z_{\text{n}} - z_{\rm U})^2},\\
\quad n = 1, 2, \dots, N,
\end{split}
\label{eq:rn_definition_split}
\end{equation}
and the near-field beamforming vector of the RIS \(\psi_n\) is used to steer the signal through \(B\) bits quantized levels  
\begin{equation}
\psi_{\text{n}} \in \Bigg\{ 
0, \frac{2\pi}{2^B}, \frac{2 \cdot 2\pi}{2^B}, \dots,\\
\frac{(2^B-1) \cdot 2\pi}{2^B} 
\Bigg\}, \quad n = 1, 2, \dots, N.
\label{eq:psi_n_levels_split}
\end{equation}

In the following sections, we present the problem formulation, explaining the rationale for utilizing both the far-field and near-field SNR models to minimize \(N\) and, consequently, reduce \(T_{\text{c}}\). We also introduce the proposed optimization algorithm, which is executed offline to ensure that its convergence time does not impact the responsiveness of the PHO framework during real-time handover operation.

\section{Problem formulation and suggested optimization  } \label{prob}
The main objective of this work is to minimize the number of allocated RIS elements \( N \), and subsequently the RIS configuration time \( T_{\text{c}} \). Striking a balance between minimizing \( N \) and preserving the required link performance is essential to ensure efficient and timely execution of the PHO process.

\subsection{Problem Formulation }
The purpose of the optimization process presented in this work is to determine the optimal configuration for the end-to-end RIS-assisted system, taking into account key parameters such as the number of dedicated subcarriers \( K \), the transmitter specifications represented by the required number of antenna elements \( M_{\text{active}} \) depending on the number of RF chains dedicated, the transmission power, and the selection of optimal beamforming vectors from the BS to the RIS and from the RIS to the blocked user all aiming to minimize the allocated number of RIS elements to establish the VLoS link. 

Obtaining a closed loop objective function form to minimize \(N\) from the near-field Eq. (\ref{eq:17near}) is not possible thus we start by finding an objective function utilizing the far-field Eq. (\ref{eq:17}) assuming that the RIS has an idealistic continuous phase shift and thus the beamforming vector $v^{\star}$ is assumed to be able to counter effect the values produced by  \(  f(\theta, \phi) \) in Eq. (\ref{eq:16}).  By using Eq. (\ref{eq:14}) to estimate  \(\beta_{\text{BI}}\), \(\beta_{\text{IU}}\)
and substituting both along with Eq. (\ref{eq:16}) into Eq. (\ref{eq:17}), \(\gamma_{\text{VLoS,far}}\)
 can be expressed as   
\begin{equation}
    \gamma_{\text{VLoS,far}} = \frac{1}{K} 
    \sum_{k=1}^{K} \rho M_{\text{active}} N^2 C 
    \left| \textbf{h}_{c,k} \textbf{f}_{\text{BI}}^{\star} \right|^2
    \label{eq:18a},
\end{equation}
Where \(C\) is 
\begin{equation}
    C = \frac{A_{\text{MIMO}} \, A^2 \, A_{\text{U}}}
    {(\lambda)^4 \, (d_{\text{BI}} d_{\text{IU}})^2}
    \label{eq:18b},
\end{equation}
Through mathematical manipulation of Eq. (\ref{eq:18a}) an objective function to minimize  \( N \) can be found as
\begin{equation}
\resizebox{0.5\hsize}{!}{$
\text{(P1)}: \quad 
\underset{M, K, \textbf{f}_{\text{BI}}^{\star}}{\text{min}} \quad 
N_{\text{far}} = \sqrt{
\frac{
\gamma_{\text{VLoS,far}} \, K
}{
\rho \, M_{\text{active}} \, C
\sum_{k=1}^{K} 
\left| \textbf{h}_{c,k} \, \textbf{f}_{\text{BI}}^{\star} \right|^2
\label{far}
}
}
$}
\end{equation}
 subject to:
    \begin{align}
    \gamma_{\text{VLoS,far}} &\geq \gamma_{\text{th}}, \tag{22.a} \\
    m_{\text{sub}} \leq M_{\text{active}} &\leq M_{\text{max}}, \tag{22.b}\\
    P_{\text{min}} \leq P_{\text{t}} &\leq P_{\text{budget}}, \tag{22.c} \\
    1 \leq K &\leq K_{\text{max}}, \tag{22.d}
    \end{align}
Further details of the optimization problem derivation are provided in the supplementary material (Section~S.I). This objective function aims to minimize \(N\) through the far-field channel model by exploring the optimal combination of parameters required to establish the VLoS connection. These parameters include the number of transmitter antennas, ranging from a single subarray \(m_{\text{sub}}\) up to utilizing all subarrays \(M_{\text{max}}\) by activating all available RF chains, as well as the dedicated subcarriers \(K\) and other system constraints.

The far-field optimization is performed first because it provides a closed-form objective function that directly minimizes the required number of RIS elements. In contrast, the near-field model in Eq.~(\ref{eq:17near}) requires joint optimization of the RIS phase configuration and transmit beamforming vectors, resulting in significantly higher computational complexity. After obtaining the optimized values \(N_{\text{far}}\), \(K_{\text{far}}\), \(M_{\text{far}}\), and the associated system parameters, they are substituted into Eq.~(\ref{eq:17near}) to evaluate \(\gamma_{\text{VLoS,near}}\).

At this stage, two outcomes are possible. If \(\gamma_{\text{VLoS,near}} \geq \gamma_{\text{th}}\), the target SNR requirement is satisfied and no further optimization is required. Otherwise, if \(\gamma_{\text{VLoS,near}} < \gamma_{\text{th}}\), the optimized values \(N_{\text{far}}\), \(K_{\text{far}}\), and \(M_{\text{far}}\) are used as lower bounds for the near-field optimization. Additional resources such as increasing \(K\), \(M_{\text{active}}\), and, if necessary, the number of RIS elements are increased as a last resort to achieve the traget $SNR$ for the near-field after exhausting all available resources. Based on these lower bounds, we formulate the near-field optimization problem, which aims to maximize the VLoS SNR:
\begin{equation}
\resizebox{0.35\hsize}{!}{$
\text{(P2)}: \quad 
\underset{M, K, \mathbf f_{\mathrm{BI}}, \{\psi_n\}}{\text{max}} \quad 
\gamma_{\mathrm{VLoS,near}}
$}
\label{near}
\end{equation}
subject to:
    \begin{gather}
    M_{\text{far}} \leq M_{\text{active}} \leq M_{\text{max}}, \tag{23.a} \\
    P_{\text{far}} \leq P_{\text{t}} \leq P_{\text{budget}}, \tag{23.b} \\
    K_{\text{far}} \leq K \leq K_{\text{max}}, \tag{23.c} \\
    N_{\text{far}} \leq N_{\text{near}} \leq N_{\text{RIS,full}}, \tag{23.d} \\
    \psi_{\text{n}} \in \Big\{ 0, \frac{2\pi}{2^B}, \frac{2 \cdot 2\pi}{2^B}, \dots, \frac{(2^B-1) \cdot 2\pi}{2^B} \Big\}. \tag{23.e}
\end{gather}
After finding the optimal $N_{\text{near}}$ that achieves the target $\gamma_{\text{VLoS,near}}$, we use the allocated number of RIS elements to calculate $T_{\text{c}}$ from Eq. (\ref{eq:1}).  
The two-stage optimization is important. In the first stage, we find the values under ideal far-field conditions using Eq. (\ref{far}) and then use the obtained values as a lower bound for Eq. (\ref{near}). This approach helps reduce computational complexity by narrowing the solution space defined by the constraints of Eq. (\ref{near}), since the complexity is highly dependent on factors such as $M$, $N$, and the relative beamforming codebooks. In mmWave systems, these factors can be on the order of hundreds or even thousands. Further details are provided in the computational complexity section.

The objective functions and the associated constraints are designed to minimize the configuration time for establishing an RIS-assisted link by indirectly reducing the required number of  allocated RIS elements needed to achieve a target SNR. The constraints enforce practical limitations, such as meeting minimum SNR requirements, respecting hardware and power limitations, and selecting beams from discrete codebooks that best align with the propagation environment.
The optimization problems formulated in equations \eqref{far} and \eqref{near}  are both non-convex because of variables such as $M$, $K$ and $f_{\text{BI}}^{\star}$ are discrete in nature, while others like $\rho$ can be considered continuous, resulting in a mixed-variable domain. Also, the beam selection functions to find optimal BS beam $f_{\text{BI}}^{\star}$ presented in Eq. (\ref{eq:11_snr})  is non-convex rendering traditional gradient-based or convex solvers unsuitable for this scenario.
\subsection{Proposed optimization algorithm} 
This work is the first to explore the concept of environment-aware RIS-assisted PHO, establishing a foundational baseline for future research directions such as proactive user tracking to address the limitations associated with the passive nature of RIS technology. The optimization is performed in an offline approach to construct a site-specific CKM, thereby ensuring that the algorithm's convergence time does not impact real-time system responsiveness. The CKM enables a fast RIS configuration setup selection to steer the signal toward the predicted blockage location in online scenarios. This approach enables efficient and timely handover execution. Future research may involve periodic updates to the CKM using ML prediction techniques and digital twin models to further enhance accuracy in highly dynamic mmWave environments.

Metaheuristic optimization methods have become a natural choice for effectively handling non-convexity and mixed-variable system models. These algorithms do not rely on gradient information and can explore large and complex solution spaces for suboptimal solutions. Among various metaheuristic methods such as genetic algorithms (GA), simulated annealing (SA), and ant colony optimization (ACO), particle swarm optimization (PSO) stands out as particularly well-suited for this problem. 
PSO offers several advantages in this context. It is simple to implement, requiring fewer hyperparameters than GA or SA, and due to this fact it exhibits faster convergence time per iteration. Moreover, PSO leverages swarm intelligence by balancing individual particle exploration and global exploitation, which improves the search for the best suboptimal solution. 
PSO is an effective optimization algorithm widely employed for solving complex problems involving multiple parameters and of a discrete nature.
The use of PSO has been applied to optimize the states of RIS elements for uniform phase changes~\cite{[17-2]},~\cite{[18]}, and to optimize the spacing of RIS elements~\cite{[3-18]},~\cite{[17-3]}. In our previous work~\cite{[23]}, we used PSO to efficiently partition an extremely large RIS (XL-RIS) into subarrays to support multiple users without compromising link quality.
In this work, we propose the use of PSO to achieve the desired trade-off between minimizing the number of RIS elements while maintaining the desired link SNR, leading to estimating the RIS configuration time \(T_{\text{c}}\). The optimizer identifies the optimal parameters, such as \(K\), \(P_{\text{t}}\), \(M_{\text{active}}\), and the optimal beam \( \textbf{f}^{\star}_{\text{BI}} \) by following the steps outlined in (Algorithm 1) and
the proposed algorithm is used to solve (\(\mathbf{P1}\)) and then update the solution space to solve (\(\mathbf{P2}\)) reducing computational complexity. A detailed PSO sensitivity analysis is provided in the supplementary material Section~S.II.

The algorithm begins by setting the maximum number of iterations and the number of particles that will explore both the local and global solutions within the solution space of the parameters. The first step is to find the optimal transmitter activated subarrays specifications represented by \(M_{\text{active}}\). 
At the start of each iteration, the algorithm initializes the population of particles at random positions to search in parallel across the entire solution space. The particles randomly select \( M_{\text{active}} \), which represents the required number of UPA transmitter elements needed to establish a link with the desired SNR. The number of transmitter antenna elements, $M_{\text{active}}$, is selected from a range bounded by $m_{\text{sub}}$ and $M_{\text{max}}$. Within this range, $M_{\text{active}}$ is discretized into fixed block sizes of 16, 32, 64, and so on, up to $M_{\text{max}}$. For example, if the maximum number of available transmit antennas is 64, the possible values for $M_{\text{active}}$ would be 16, 32, and 64, depending on the number of activated RF chains. One activated RF chain corresponds to the smallest MIMO subarray of $m_{\text{sub}} = 16$ antenna elements, two RF chains yield 32 elements, and so on.
In this work, the optimization process is conducted for three different values of \( M_{\text{max}} \), specifically 128, 256, and 512, each offering a distinct set of discrete options for \( M_{\text{active}} \). These variations allow for evaluating the impact of transmitter array size on system performance and RIS-assisted link efficiency.

\begin{algorithm}[h]
    \caption{PSO for minimizing allocated RIS elements and RIS configuration time.}
    \small
    \begin{algorithmic}[1]
        \State Choose population size \textit{num\textunderscore particles}, number of iterations, inertia weight \textit{w=0.7}, cognitive coefficient \textit{c1=1.5}, and social coefficient \textit{c2=1.5} to solve (\(\mathbf{P1}\)).
        \State Initialize the subcarriers values with $K=1$ and set the value of \( K_{\text{max}} \).
        \State Set the threshold value of RIS elements \(N_{\text{RIS,full}}\) when exceeded $K$ is incremented.
        \State Set the minimum single RF chain $m_{\text{sub}}$ and maximum $M_{\max}$ values for \(M_{\text{active}}\).
        \State Initialize particle positions \(M_{\text{active}}\) as random multiples of the block size within [$m_{\text{sub}}, M_{\max}$].
        \State Initialize \textit{$P_{\text{t}}$} within [\textit{P\textunderscore min}, \textit{P\textunderscore max}].

        \For{iter = 1 to \textit{num\textunderscore iterations}}
            \For{each particle}
                \State Evaluate $\mathbf{f}_{\text{BI}}$.
                \State Evaluate \(N\).
                \If{\textit{N} exceeds \(N_{\text{RIS,full}}\)}
                    \State Increment \textit{K}.
                \EndIf
                \If{current $N$ is better than local best}
                    \State Update local best values.
                \EndIf
                \If{current $N$ is better than global best}
                    \State Update global best values.
                \EndIf
            \EndFor
            \For{each particle}
                \State Update velocities using PSO equations.
                \State Update particle positions.
                \If{\textit{K} exceeds $K_{\text{max}}$}
                    \State Set \textit{K} to $K_{\text{max}}$.
                \EndIf
            \EndFor
        \EndFor
        \State Initialize PSO for problem (\(\mathbf{P2}\)).
        \State \Return $N_{\text{near}}$, $T_{\text{c}}$, $M_{\text{active}}$, and $K_{\text{near}}$.
    \end{algorithmic}
\end{algorithm}

To define the transmit power \( P_{\text{t}} \) utilized by the transmitter, we focus on optimizing \( \rho = \frac{P_{\text{t}}}{\sigma^2} \). Various values of \( \rho \) have been considered in the literature, such as \( \rho = 5 \) dB in \cite{[3-6]}. In this study, we assume that the minimum transmitter configuration, \( m_{\text{sub}} \), corresponding to a 16-element single RF chain antenna block, achieves a baseline of \( \rho = 0 \) dB. As the algorithm increases the number of transmitter elements to the next block size (e.g., 2 RF chains \( M_{\text{active}} = 32 \)), the available transmit power budget is adjusted to improve \( \rho \), allowing values from 0 dB up to 3 dB. More generally, each doubling of \( M_{\text{active}} \) results in a 3 dB increase. This strategy aligns with the physical interpretation that increasing the array size enhances directional gain, thereby improving \( \rho \). Accordingly, the optimization algorithm evaluates all possible combinations of \( M_{\text{active}} \) and its transmit power, ultimately selecting the configuration that provides the most favorable trade-off between link quality and power consumption. The next step involves calculating the best beamforming codebook vector, \( \textbf{f}_{\text{BI}} \), used for transmission from BS to RIS. These calculations are based on the randomly selected \( M_{\text{active}} \) positions. 

The initial value of the number of subcarriers, \( K \), is set to 1 and is incremented cautiously due to the associated computational complexity. To constrain the system complexity and maintain practical feasibility, a threshold is introduced, limiting the maximum number of RIS elements \(N_{\text{RIS,full}}\) to 1089 arranged in a \( 33 \times 33 \) grid to maintain the odd side dimension ensuring a fixed RIS center across all allocations as introduced in the system model. This threshold is based on our previous findings in \cite{[3-3]}, where 1000 elements were required to establish a reliable RIS-assisted link for a PHO. In the present work, the primary objective is to reduce the required number of allocated RIS elements based on the far-field conditions \( N_{\text{far}} \) while achieving comparable performance. Once the algorithm computes \( N_{\text{far}} \) using the current values of system parameters, it compares the result against the defined threshold. If the computed value exceeds the threshold, \( K \) is incremented, and the process is repeated to recompute \( N_{\text{far}} \). This iterative procedure continues until the condition \( K \leq K_{\text{max}} \) is satisfied, ensuring that the optimal combination is selected to minimize \( N_{\text{far}} \) without exceeding the maximum permissible complexity.
At this stage, the best personal (local) value of \( N_{\text{far}} \) is updated based on the minimum value across all particles. The global best position is also updated to achieve the minimal \( N_{\text{far}} \) across all iterations, which is then returned as the optimized \( N_{\text{far}} \). Additionally, the algorithm determines the optimal set of parameters required to achieve the desired SNR at the optimized \( N_{\text{far}} \). These include the required transmitter array \( M_{\text{active}} \), the optimal BS beam \( \textbf{f}_{\text{BI}}^{\star} \) transmitting to the RIS, and the best number of subcarriers \(K_{\text{far}}\) allocated for the user. 
These values are then used to evaluate \(\gamma_{\text{VLoS,near}}\).
If \(\gamma_{\text{VLoS,near}} \geq \gamma_{\text{th}}\), the target SNR requirement is satisfied otherwise, if \(\gamma_{\text{VLoS,near}} < \gamma_{\text{th}}\), the far-field values are used as lower bounds in the constraints of (\(\mathbf{P2}\)) to evaluate \(\gamma_{\text{VLoS,near}}\) to extract the optimal \(N_{\text{near}}\), which is then used to estimate \(T_{\text{c}}\). The complete flowchart of the proposed optimization framework is provided in the supplementary material Section~S.III.

\subsection{Computational complexity}
The computational complexity associated with  \(N\) and $T_{\text{c}}$ depends on the considered scenario. In the far-field continuous phase shift, the RIS beamforming vector is analytically derived to counteract the angle-of-arrival and angle-of-departure components, eliminating the need for RIS optimal beam search. The complexity in this case primarily arises from evaluating the achievable SNR over a finite set of transmit-side beams \( \mathcal{F} \) which is dependent on \(M_{\text{active}}\), resulting in a complexity on the order of \( \mathcal{O}(I\cdot P\cdot M_{\text{active}} \cdot K) \), where \( K \) is the number of allocated subcarriers to a blocked user and where \(I\) and  \(P\) denote the number of PSO iterations, particles respectively. 

In contrast, the proposed near-field optimization requires joint adaptation of the RIS phase configuration and transmit beamforming vectors across all subcarriers. The overall computational complexity is dominated by the beamforming and SNR evaluations within the PSO process, scaling as 
\(\mathcal{O}(I \cdot P\cdot K \cdot (M_{\text{active}} + N))\), 
 Consequently, the complexity increases linearly with the number of transmit antennas \(M_{\text{active}}\), RIS elements \(N\), and subcarriers \(K\). While this heuristic approach significantly reduces the need for exhaustive beam search over large codebooks, it remains computationally more demanding than the continuous (ideal) beamforming case due to iterative optimization over a high-dimensional joint search space. To reduce the computational burden of joint near-field optimization, we adopt a two-stage warm-start strategy. First, a far-field model is optimized to obtain a coarse solution \((K_{\mathrm{far}},M_{\mathrm{far}},N_{\mathrm{far}})\). These values are then used to restrict the near-field search space and to initialize  PSO particles around the far-field solution as this pruning reduces the number of PSO evaluations.

\section{Suggested RIS-Assisted PHO Framework} \label{VAWC}
In this work, the bus in Fig. \ref{fig:4} prevents UPA sector 1 from establishing a LoS link to the user. When this blockage event is detected by the VAWC framework integrated with the BS, it triggers a PHO to UPA sector 2's RIS-assisted link. The new RIS-assisted link will be selected from a subset of VLoS beam indices stored at the BS as part of the RIS-assisted CKM database.
The PSO algorithm determines the optimal $M_{\text{active}}$ used in UPA sector 2 and the optimal allocated $N_{\text{near}}$ for the RIS required to establish the link to the blocked area based on the desired SNR. Additionally, the RIS-assisted CKM contains the optimal parameters, calculated based on the outcome of the PSO algorithm performed using an offline estimation approach, to enable a fast response during online scenarios. These parameters include the optimal beams \(\textbf{f}_{\text{BI}}^{\star} \), and the corresponding transmitted power to the blocked location. Finally, the PSO algorithm will determines \(T_{\text{c}}\) based on the optimized value of \(N_{\text{near}}\), and \(T_{\text{c}}\) will be incorporated as an additional time to execute the updated PHO timing requirements, as shown in Eq. (\ref{eq:5}), ensuring a smooth and accurate timing to trigger the PHO procedure.

The proposed framework can be further extended to support more practical and large-scale deployment scenarios. First, to support multiple users, modifications to the physical access system are required. Since the considered MIMO sector already utilizes hybrid beamforming, the BS can simultaneously project multiple beams depending on the available RF chain configuration. For the RIS-assisted link, multiple RISs may be deployed, where each RIS serves a different user or coverage region. Alternatively, an XL-RIS \cite{[23]} can be partitioned into multiple tiles, where each tile independently serves a different user. Such RISs can be mounted on multiple buildings or infrastructure points to provide extended street-level coverage.
At the control layer, the VAWCS can be further trained to include multiple blockage scenarios, such as different blockage locations, varying obstacle mobility patterns, and different UE mobility speeds. This would further enrich the CKM database to accommodate a wider range of environmental and mobility conditions.
\section{PERFORMANCE EVALUATION AND RESULTS} \label{results}
The work in this paper adopts the simulation environment described in \cite{[3-3]}, where the VAWC framework detects blockage events, user locations, and the obstacle. In this simulation, we focused on a section of the street experiencing signal blockage, specifically between \( x = 15 \) and \( x = 22 \) meters. The RIS is dedicated to providing coverage for this segment of the street. Additional details are provided in the supplementary material Section~S.IV.
The proposed algorithm computes the optimal values of \( N_{\text{near}} \) and \( T_{\text{c}} \). Accordingly, the optimization process entails identifying the optimal number of RIS elements may not always correspond to a perfect square. In such cases, the algorithm approximates the result to the nearest perfect square with odd number in each side to facilitate practical deployment.

\begin{figure*}[t]
    \centering
    \begin{subfigure}[b]{0.43\textwidth}
        \centering
        \includegraphics[width=\textwidth]{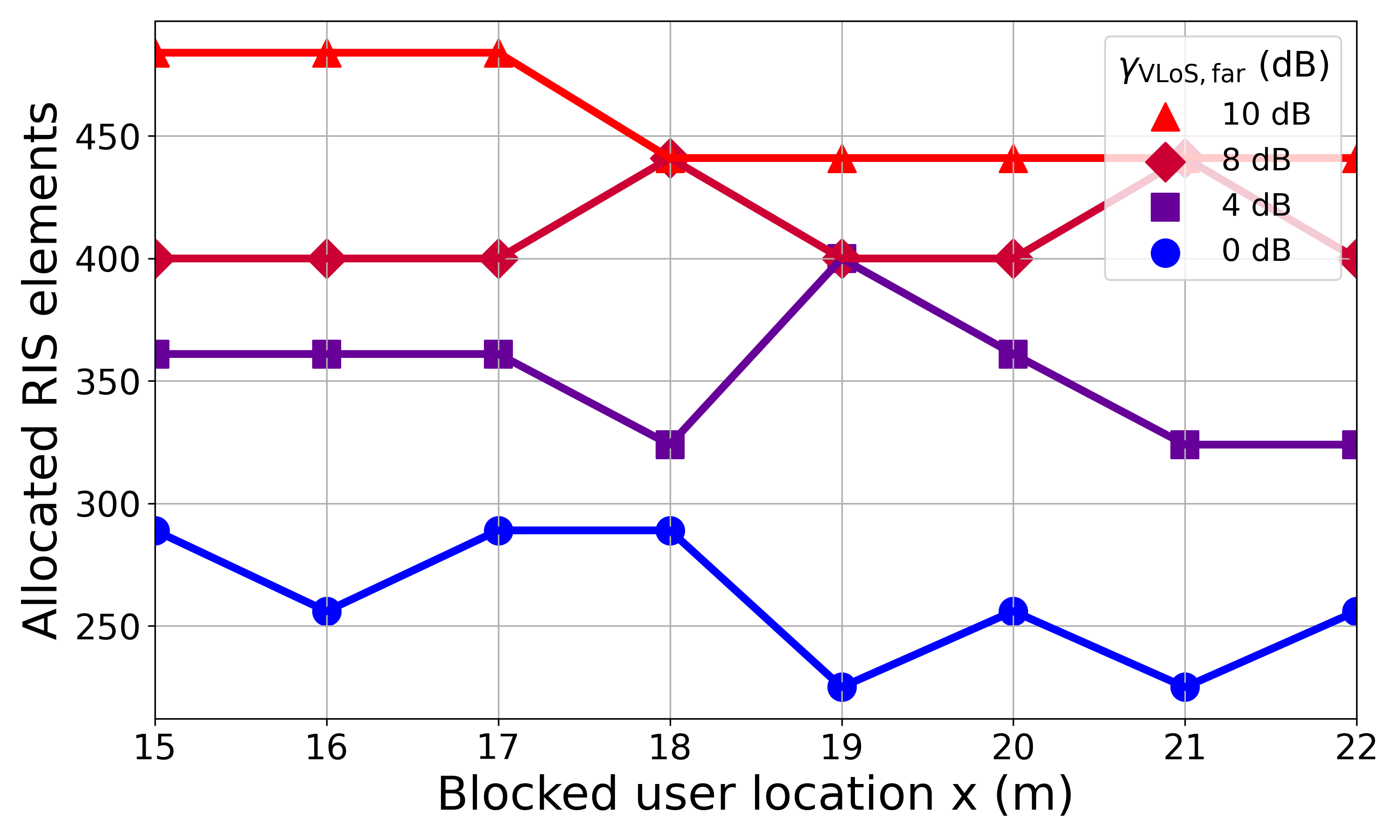}
        \caption{Effect of RIS area on SNR}
        \label{fig3-2}
    \end{subfigure}
    \hfill
    \begin{subfigure}[b]{0.43\textwidth}
        \centering
        \includegraphics[width=\textwidth]{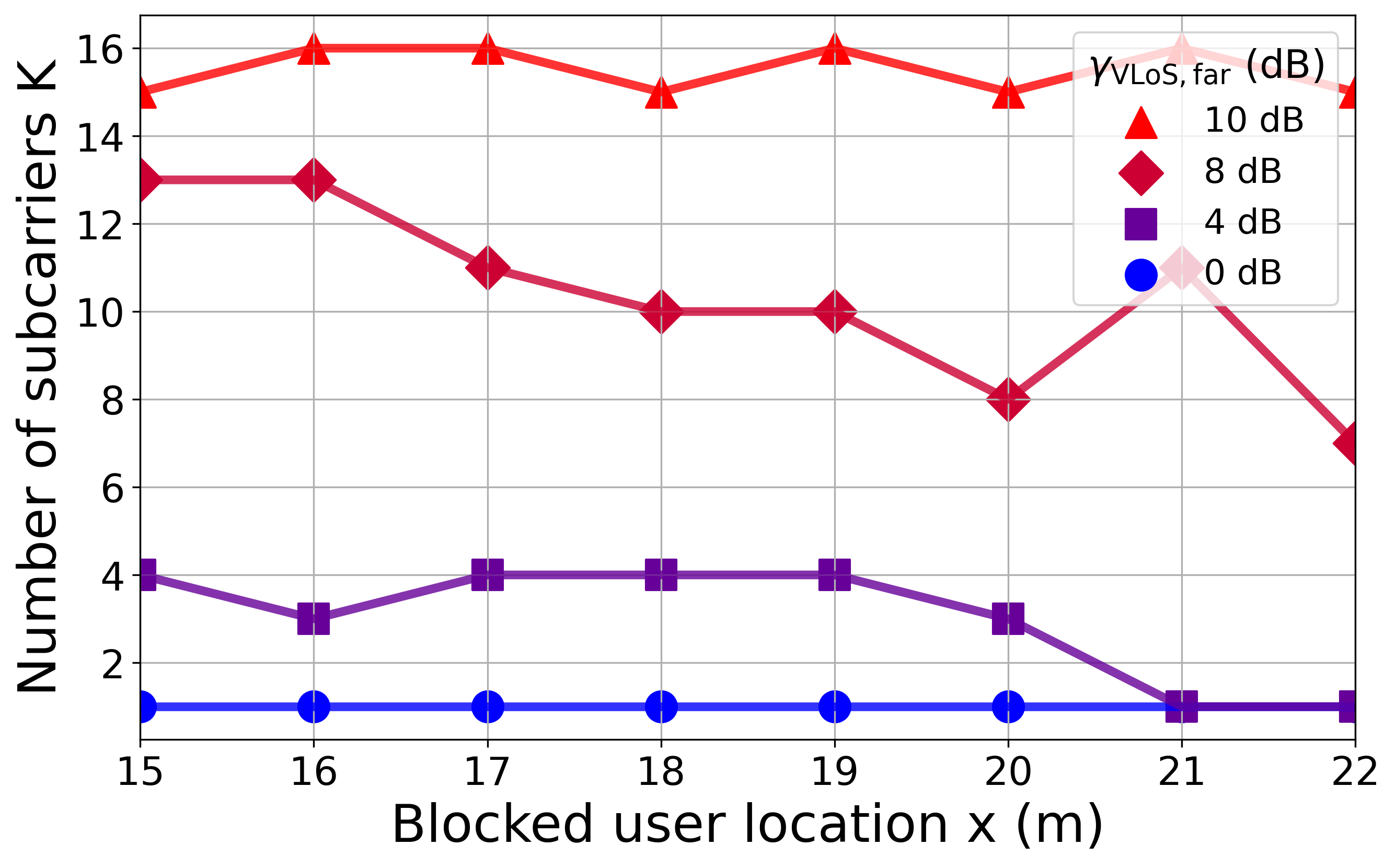}
        \caption{Effect of subcarriers on SNR}
        \label{fig3-3}
    \end{subfigure}
    \caption{Comparative plots of RIS area and subcarrier effects on SNR ( $M_{\text{active}}=256$) far-field continuous phase shift.}
    \label{fig:combined}
\end{figure*}

The results presented in Fig.~\ref{fig:combined} illustrate the required RIS area and the corresponding subcarrier allocation across various segments of the blocked region in the far-field ideal phase-shift model. Fig.~\ref{fig3-2} depicts the variation of the allocated RIS elements with respect to different SNR targets. The estimated square RIS areas range from \( N = 289 \) elements, arranged in a \( 17 \times 17 \) grid for the lowest SNR of \( \gamma_{\text{VLoS,far}} = 0 \), up to \( N = 484 \) elements, arranged in a \( 22 \times 22 \) grid for the highest rate of \( \gamma_{\text{VLoS,far}} = 10 \). It is worth mentioning that, in the first stage (far-field) of the optimization, the algorithm does not approximate the required allocation of RIS elements to an odd-sided square area; this approximation is performed in the near-field optimization stage.
It is also worth mentioning that when the blocked user is closer to the BS located at \(x = 45\), fewer RIS elements and subcarriers are required. This is clearly demonstrated for a user at \(x = 15\) requires more resources than the user at \(x = 22\), as the user at  \(x = 15\) is located farther from the BS.
Fig.~\ref{fig3-3} presents the associated subcarrier allocations required for each RIS configuration to maintain the same target SNR. For \( \gamma_{\text{VLoS,far}} = 0~\mathrm{dB} \), the number of allocated subcarriers is \( K = 1 \), whereas for \( \gamma_{\text{VLoS,far}} = 10~\mathrm{dB} \), the required number of subcarriers reaches \( K = 16 \). The proposed algorithm was able to achieve higher SNR with the same RIS area by increasing the number of allocated subcarriers. This is reflected in the overlap at \( x = 18 \) of the \( \gamma_{\text{VLoS,far}} = 8~\mathrm{dB} \) and \( \gamma_{\text{VLoS,far}} = 10~\mathrm{dB} \) curves in Fig.~\ref{fig3-2}, both corresponding to allocated RIS elements of \( N = 441 \), but with allocated subcarriers of \( K = 10 \) and \( K = 15 \), respectively, as shown in Fig.~\ref{fig3-3}.
These findings highlight an important trade-off: the allocation of additional subcarriers can compensate for increasing allocated RIS elements to achieve higher SNR, and vice versa. This enables the system to flexibly balance between resource allocation requirements and signal processing complexity. Consequently, once the RIS area is fixed to meet configuration time constraints, dynamic subcarrier allocation can be employed to achieve the required link performance based on user demands, without requiring reallocation or modifications to the RIS structure.
\begin{figure}
\centering
\includegraphics[width=0.44\textwidth]{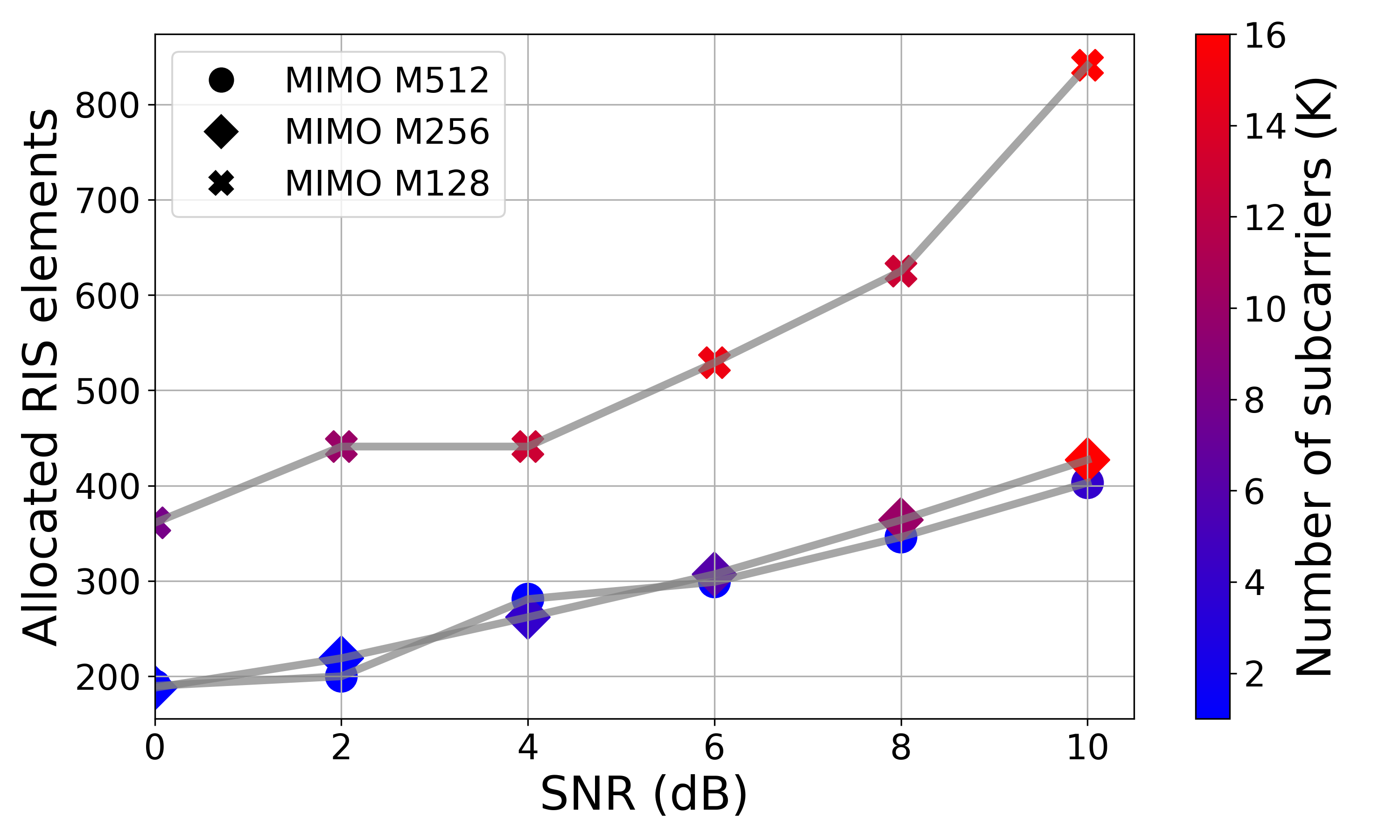}
 \caption{Effect of transmitter MIMO on SNR, RIS area, and subcarriers ( far-field continuous phase shift scenario).}
 \label{fig3-4}
\end{figure}

The results in Fig. \ref{fig3-4} highlight another factor that contributes to minimizing the RIS area: the use of narrower beams projected onto the RIS by increasing the number of transmitter MIMO elements. This figure illustrates the correlation between the number of subcarriers \( K \), the number of transmitter MIMO elements \( M_{\text{active}} \), and the required RIS area for different values of SNR. In general, increasing the number of MIMO elements enhances performance. A comparison of \( M_{\text{active}} = 128 \), \( M_{\text{active}} = 256 \), and \( M_{\text{active}} = 512 \) shows that a transmitter with \( M_{\text{active}} = 512 \) requires a smaller RIS area to achieve the same $SNR$ compared to \( M_{\text{active}} = 256\), particularly noticeable at SNR\( > 6 \). Another perspective on the impact of the transmitter MIMO can be observed for  SNR\( < 6 \), where both \( M = 512 \) and \( M = 256 \) share the almost the same RIS area. However, the \( M = 512 \) transmitter requires 1 to 4 subcarriers to achieve the required SNR, whereas the \( M = 256 \) transmitter requires 4 to 8 subcarriers. Overall, larger MIMO arrays consistently deliver better performance while reducing the need of other resources, either by reducing the required number of subcarriers or by minimizing the RIS area.
\begin{figure*}[t]
    \centering

    \begin{subfigure}[b]{0.32\textwidth}
        \centering
        \includegraphics[width=\textwidth]{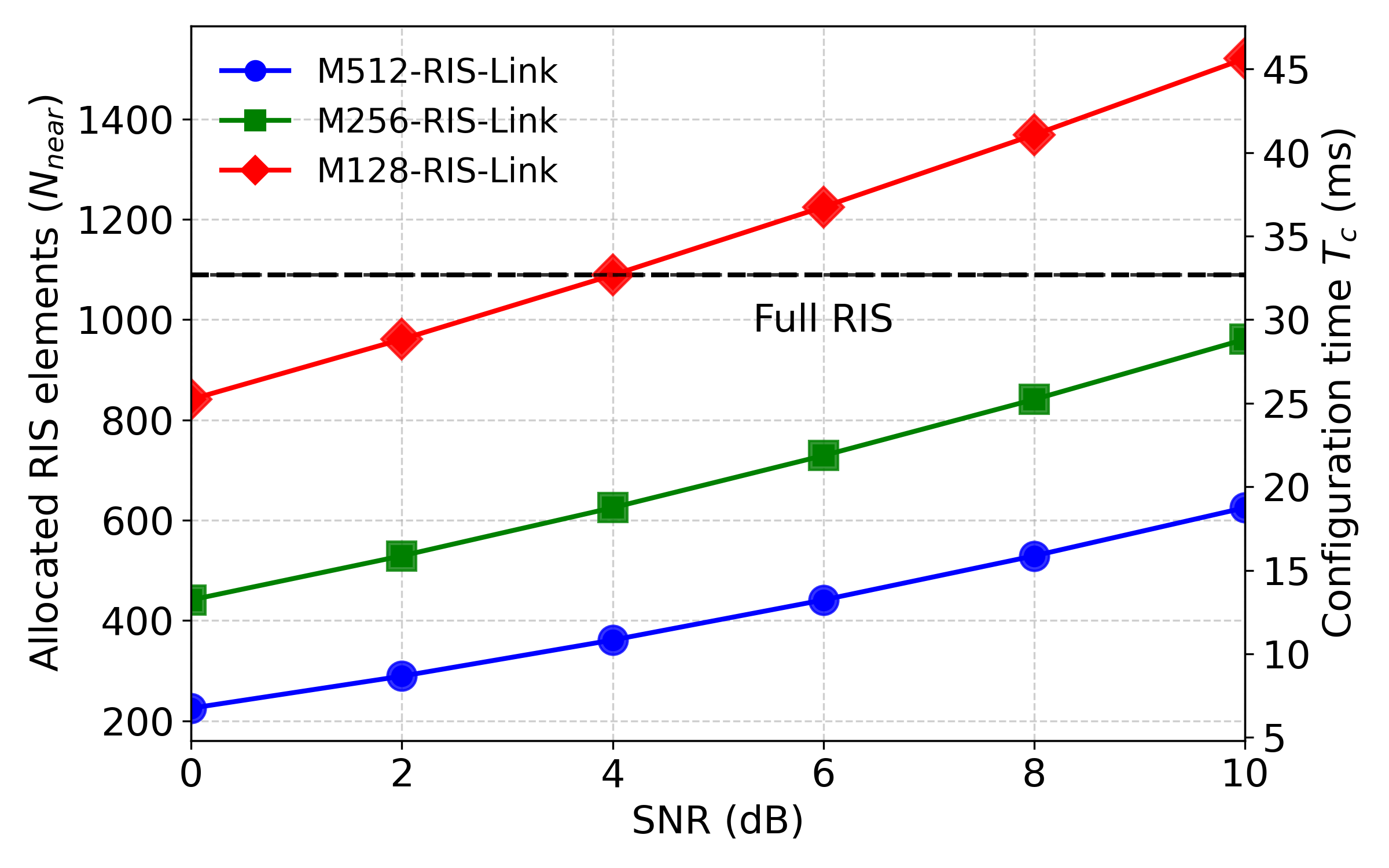}
        \caption{RIS configuration time ($B=3$).}
        \label{fig3-5}
    \end{subfigure}
    \hfill
    \begin{subfigure}[b]{0.32\textwidth}
        \centering
        \includegraphics[width=\textwidth]{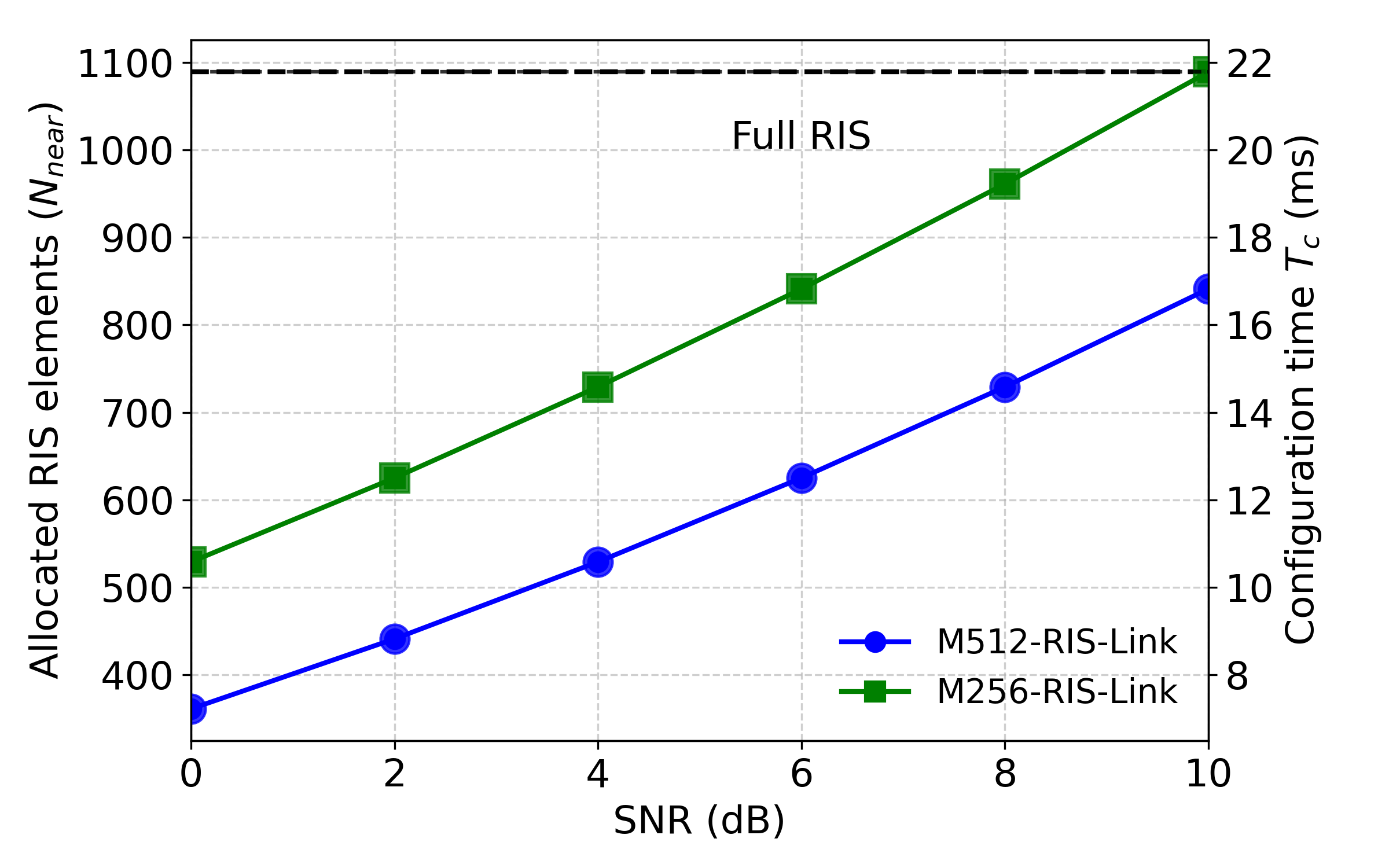}
        \caption{RIS configuration time ($B=2$).}
        \label{fig3-6}
    \end{subfigure}
    \hfill
    \begin{subfigure}[b]{0.32\textwidth}
        \centering
        \includegraphics[width=\textwidth]{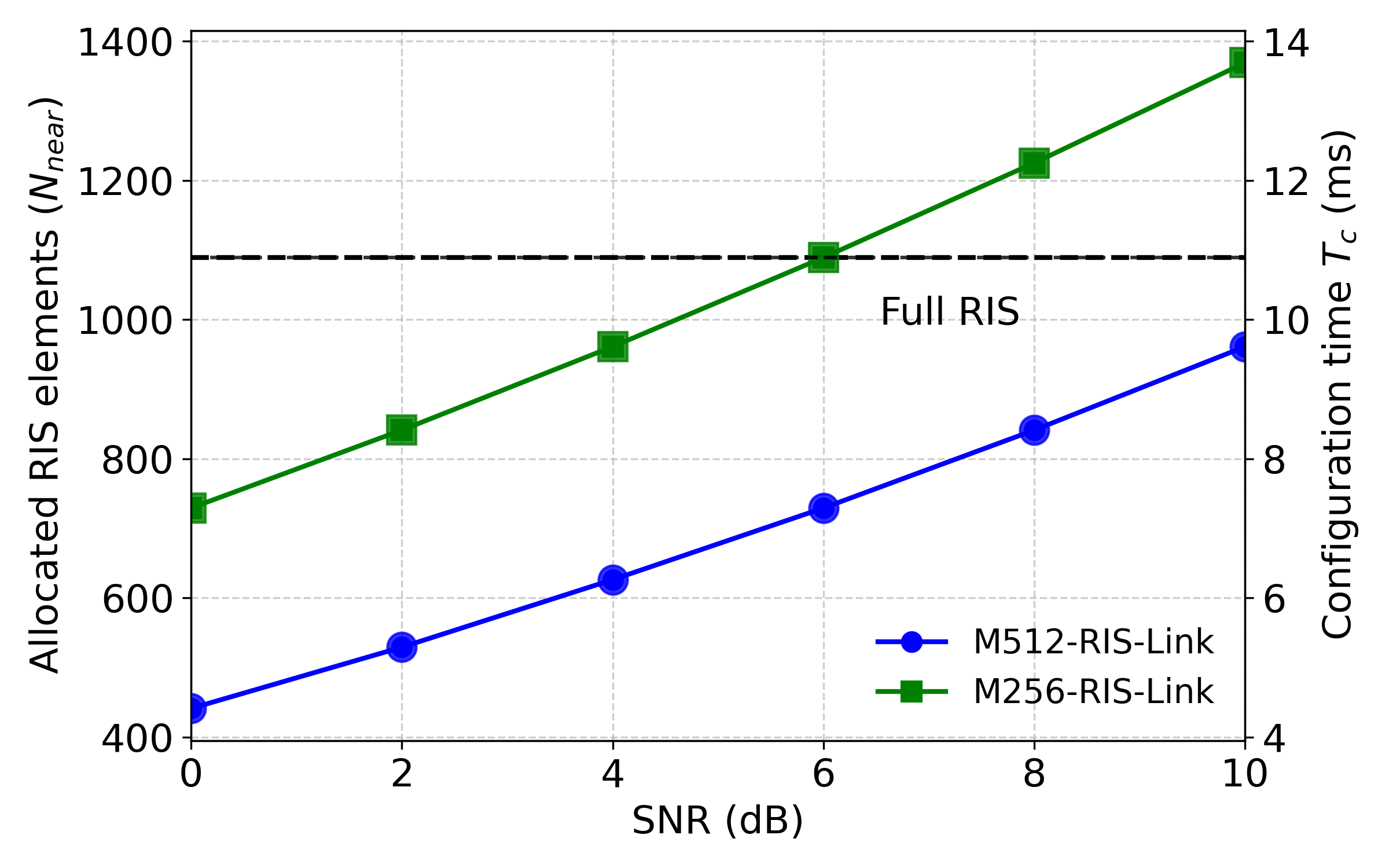}
        \caption{RIS configuration time ($B=1$).}
        \label{fig3-6c}
    \end{subfigure}

    \caption{Comparative plots of SNR, allocated RIS elements $N_{\text{near}}$ and configuration time.}
    \label{fig:combined2}
\end{figure*}

Fig.~\ref{fig:combined2} illustrates the relationship between the RIS configuration time \( T_{\text{c}} \) and the performance of the RIS-assisted link, evaluated in terms of the target SNR in the near-field discrete phase shift model and the number of allocated RIS elements required to attain that performance level.
Fig.~\ref{fig3-5} presents the performance results for discrete phase shifts with quantization levels of \( B = 3 \) bits under different \( M_{\text{active}} \) transmitter configurations and varying target SNRs, compared to full RIS elements allocation without minimization, utilizing \( N_{\text{RIS,full}} = 1089 \) elements. For the lowest (SNR\( = 0~dB \)) and \( M_{\text{active}} = 512 \), an allocation of \( N_{\text{near}} = 225 \) RIS elements, arranged in a \( 15 \times 15 \) grid, is sufficient, resulting in a configuration time of approximately \( T_{\text{c}} = 6 \, \text{ms} \). Conversely, to achieve the highest (SNR \( = 10~dB \)), the required RIS area increases to \( N = 625 \) elements, yielding a configuration time of approximately \( T_{\text{c}} = 20 \, \text{ms} \). 
For \( M_{\text{active}} = 256 \), the RIS configuration time \( T_{\text{c}} \) is higher because this transmitter setup requires a larger number of RIS elements to reach the target SNR. The \( M_{\text{active}} = 128 \) configuration was unable to achieve \( \text{SNR} > 4 \, \text{dB} \) even when all RIS elements were allocated, and it required an even greater number of elements at lower quantization levels (\( B = 1, 2 \)). Similarly, the results in Fig.~\ref{fig3-6} and Fig.~\ref{fig3-6c} present the RIS configuration times for different MIMO transmitters with quantization levels of \( B = 2 \) and \( B = 1 \), respectively. With these lower quantization levels, shorter configuration times are achieved despite requiring more allocated RIS elements compared to \( B = 3 \). However, for \( M_{\text{active}} = 256 \), the system could not achieve \( \text{SNR} > 6 \, \text{dB} \) with an RIS of \( B = 1 \), even when all available elements were allocated.

The results in Fig.~\ref{fig:combined2} underscore the importance of minimizing the RIS area to achieve faster configuration responses, which is critical for latency-sensitive applications such as autonomous vehicular communication. Hence, optimizing the RIS configuration to use the smallest feasible number of elements becomes essential in such scenarios. However, the key factor that contributes to reducing the RIS configuration time \(T_{\text{c}}\) is the number of diodes per RIS element \(B\), as it reduces the codeword size transmitted by the RIS controller to obtain the desired configuration. 

\begin{figure}[t]
    \centering
    \begin{subfigure}[b]{0.48\textwidth}
        \centering
        \includegraphics[width=\textwidth]{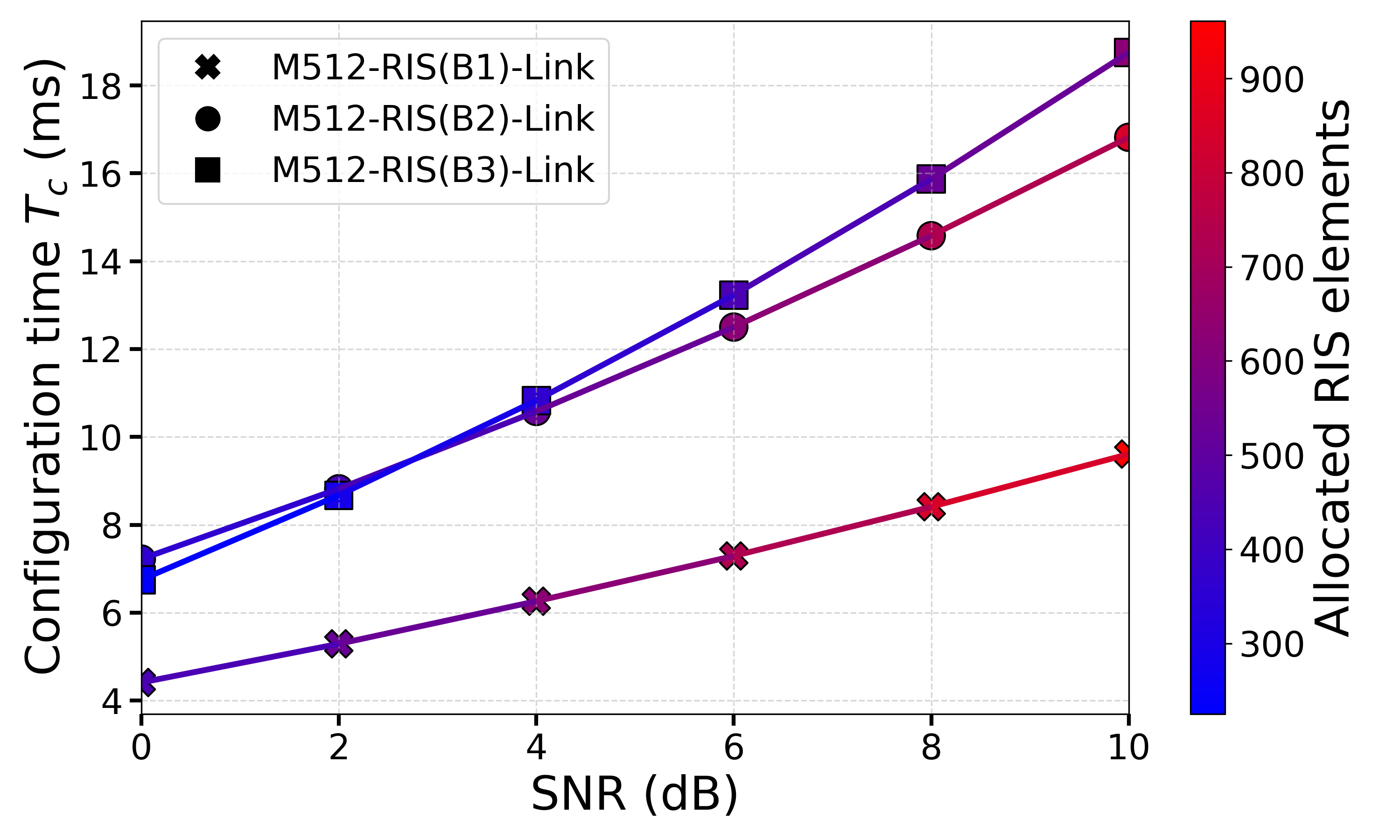}
        \caption{Effect of quantization levels on $T_{\text{c}}$ and RIS allocated elements.}
        \label{fig33}
    \end{subfigure}
    \hfill
    \begin{subfigure}[b]{0.48\textwidth}
        \centering
        \includegraphics[width=\textwidth, height=0.215\textheight]{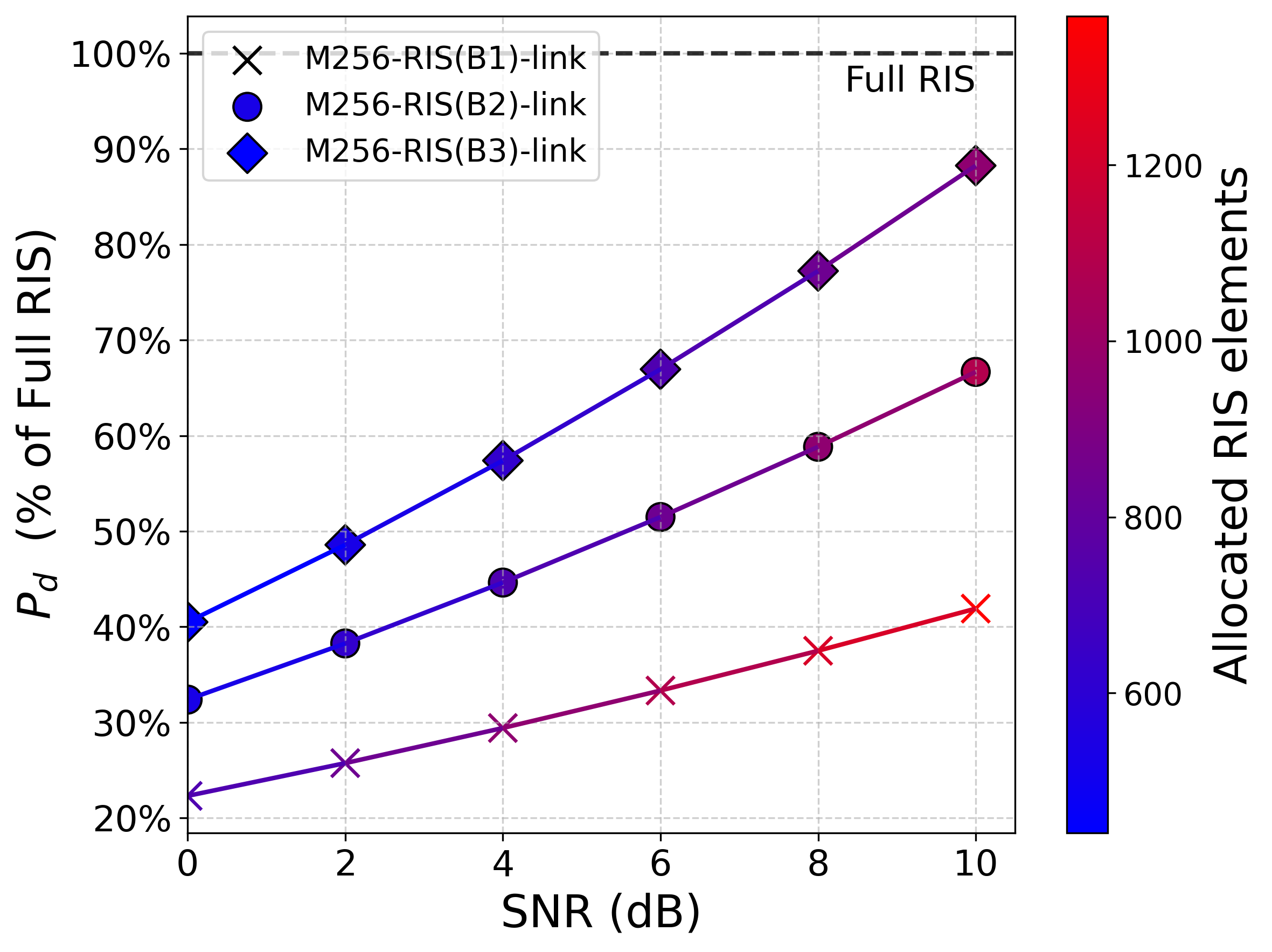}
        \caption{Effect of quantization levels on $P_{d}$ and RIS allocated elements.}
        \label{fig34}
    \end{subfigure}
    \caption{ Trade-off between link quality (SNR), allocated RIS elements, quantization levels, and their impact on  $T_{\text{c}}$ and $P_{d}$.}
\end{figure}

RISs fabricated with lower values of $B$ result in faster RIS configuration responses, as shown in Fig.~\ref{fig33}. 
The link with $M_{\text{active}} = 512$ achieves a lower configuration time when the RIS has $B = 1$, compared to 
$B = 2$ or $B = 3$. Following the same trend, the results in Fig.~\ref{fig34} also highlight that using lower 
quantization levels can enhance RIS performance in terms of energy efficiency. Specifically, the link with 
$M_{\text{active}} = 256$ exhibits lower dissipated power when the RIS uses $B = 1$ compared to higher quantization levels.
Although lower quantization levels lead to faster RIS response and improved energy efficiency, they do not reduce the 
signal processing complexity of tasks such as channel estimation or BT, since the complexity of both depends primarily 
on the number of RIS elements. An RIS with fewer quantization bits requires a larger number of elements to achieve the 
target SNR, as illustrated in Fig.~\ref{fig33}, where the M512-RIS($B1$) link achieves lower configuration time, and in 
Fig.~\ref{fig34}, where the M256-RIS($B1$) link dissipates lower power levels, yet both require more RIS elements to 
achieve the desired SNR compared to higher quantization levels.

On the other hand, increasing the MIMO array size by allocating more RF chains can enhance system performance but reduces 
the transmitter’s spatial multiplexing capability due to the increased use of subarrays to transmit to the RIS. Therefore, in 
our case, a transmitter MIMO with $M_{\text{active}} = 256$, combined with an RIS using a quantization level of $B = 3$, 
represents a feasible balance between fast RIS configuration response, lower dissipated power, and moderate transmitter 
resource allocation requirements.

Based on the findings presented above, M256-RIS link in Fig. \ref{fig3-5} with \(M_{\text{active}}=256\), the required allocated RIS elements to achieve the highest SNR is \( N_{\text{near}} = 961 \), corresponding to a configuration time of \( T_{\text{c}} = 29 \, \text{ms} \). To account for the impact of the RIS configuration time on the PHO framework discussed in \cite{[3-3]}, the original \( T_W \) value estimated in \cite{[7]} must be updated to \( T^{\text{RIS}}_{\text{W}} \), as detailed in equations (\ref{eq:4}) and (\ref{eq:5}). For instance, in \cite{[7]}, a scenario is considered where a blocked user, represented by a vehicle, moves towards the blockage area at a speed of \( 10 \, \text{miles per hour} \). The original estimation of \( T_W = 4.7 \, \text{seconds} \) was provided before triggering handover to a second BS. In our work, to ensure accurate triggering of the PHO by an RIS-assisted link instead of a second BS, the updated value \( T^{\text{RIS}}_{\text{W}} \) must be used, which accounts for the additional timing constraint introduced by the RIS configuration response. In the case for the considered M256-RIS assisted link in Fig. \ref{fig3-5} the RIS configuration time is \( T_{\text{c}} = 29 \, \text{ms} \), leading to an updated value for \( T^{\text{RIS}}_{\text{W}} = 4.67117 \, \text{s} \). This value can be used by the RIS-assisted PHO framework to trigger the handover accurately as part of the 3GPP HO preparation timing. 
Table \ref{tab:1} presents the optimal handover triggering time, accounting for the RIS configuration response, the listed RIS areas correspond to different  SNR targets, with the updated time tables the PHO algorithm can safely trigger the process with sufficient timing margin. The remaining user movement speeds and their corresponding \( T_W \) values from \cite{[7]} can be updated in a similar manner.

\begin{table}[t]
\caption{Relationship between allocated RIS elements, target SNR, and configuration times.}
\label{tab:1}
\centering
\small
\begin{tabular}{cccc}
\toprule
\makecell{\textbf{Allocated}\\\textbf{RIS elements}} & 
\textbf{Target SNR (dB)} & 
\textbf{$T_{\text{c}}$ (ms)} & 
\textbf{$T_W^\text{RIS}$ (s)} \\
\midrule
441 & 0  & 13.23 & 4.68677 \\
529 & 2  & 15.87 & 4.68413 \\
625 & 4  & 18.75 & 4.68125 \\
729 & 6  & 21.87 & 4.67813 \\
841 & 8  & 25.23 & 4.67477 \\
961 & 10 & 28.83 & 4.67117 \\
\bottomrule
\end{tabular}
\end{table}
\begin{figure}[t]
    \centering
    \begin{subfigure}[b]{0.38\textwidth}
        \centering
        \includegraphics[width=0.95\textwidth]{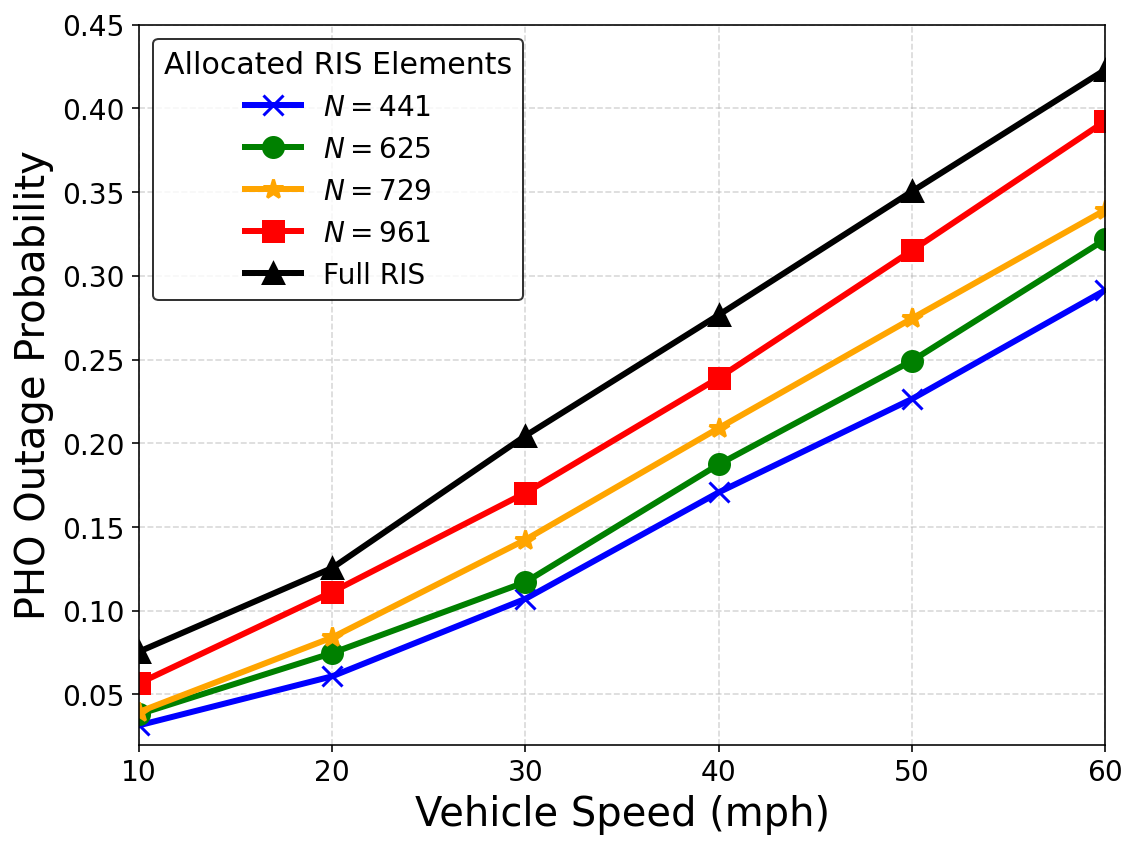}
        \caption{PHO outage probability for different RIS allocation schemes \(M_{\text{active}}=256\), \(B=3\).}
        \label{fig3-7a}
    \end{subfigure}
    \hfill
    \begin{subfigure}[b]{0.48\textwidth}
        \centering
        \includegraphics[width=\textwidth, height=0.215\textheight]{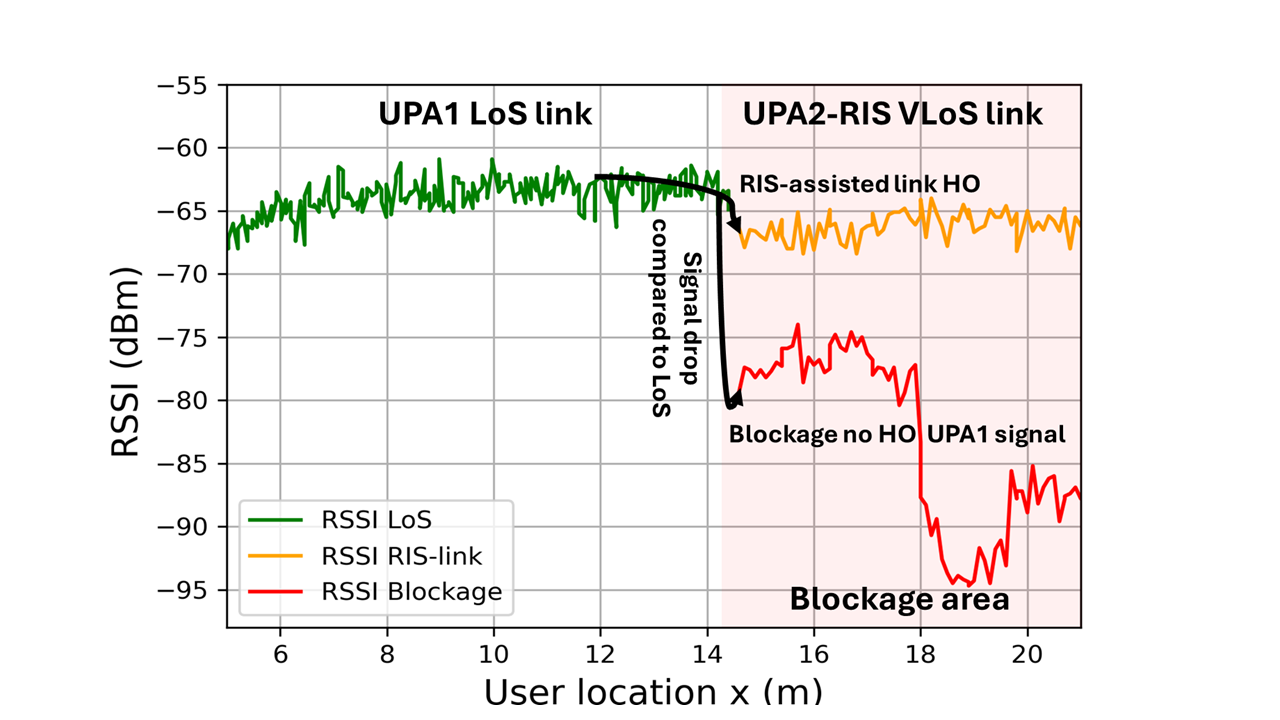}
        \caption{Comparison of RSSI for UPA1 LoS link, UPA2 RIS-assisted VLoS link, and UPA1 Blockage signal.}
        \label{fig3-7}
    \end{subfigure}
    \caption{RIS-assisted link performance: (a) PHO outage probability and (b) RSSI.}
\end{figure}
The results in Fig.~\ref{fig3-7a} present the PHO outage probability for different RIS allocations when both the vehicle and the blocking obstacle move at speeds ranging from 10~mph to 60~mph. The results show that the proposed framework consistently achieves lower outage probability while utilizing fewer RIS elements than the full RIS configuration. For example, at 60~mph, allocating \(N_{\text{near}}=441\) RIS elements reduces the outage probability from \(43\%\) to \(29.0\%\) corresponding to a reduction of $14$ percentage points. These findings highlight the importance of minimizing RIS configuration time and demonstrate the effectiveness of the proposed optimization framework under increasing mobility. 
Finally, the results in Fig.~\ref{fig3-7} present comparison in received signal strength indicator (RSSI) between three scenarios: (i) the RSSI of the LoS signal, where no blockage impedes the BS from establishing a direct link to the user using UPA1; (ii) the RSSI of the RIS-assisted link, where a beam is steered from the RIS to the blocked area using parameters optimized by the proposed PSO algorithm via a PHO to UPA2 \(M_{\text{active}}=256\) transmitter and \(RIS(B=3)\) VLoS link, denoted as the \textit{RSSI RIS-link} curve; and (iii) the UPA1 RSSI values in a blocked scenario without RIS, denoted as the \textit{RSSI Blockage} curve. While the RIS-assisted link shows reduced signal strength compared to the LoS case, it significantly outperforms the blocked scenario without RIS, improving the RSSI by approximately 15 to 30~dBm, depending on blockage severity and user location. The proposed PSO algorithm reduces the required RIS area from \( N_{\text{RIS,full}} = 1089 \), to \( N = 961 \), achieving a 12\% reduction in the number of RIS elements required to establish the link and reducing the dissipated energy by 10\% compared to the full RIS deployment as illustrated in the \(M_{\text{active}}=256\) transmitter and \(RIS(B=3)\) curve in Fig. \ref{fig34}. This reduction is attained without compromising performance and while enhancing the RSSI relative to the blocked signal scenario. Moreover, minimizing the RIS area results in faster configuration response, enabling timely beam steering toward the blocked region. By explicitly calculating and incorporating the RIS configuration time into the handover process, the PHO can be triggered more accurately, ensuring uninterrupted service for blocked users.

\section{Conclusion} \label{conc}
This work presents a novel RIS-assisted VAWC framework capable of providing seamless coverage by restoring blocked LoS links in mmWave networks. It further enhances PHO procedures by minimizing the number of allocated RIS elements to ensure faster RIS configuration response. The proposed approach facilitates accurate PHO-triggered handovers, improving mmWave signal reliability in obstructed scenarios. The optimization process identifies the optimal system configuration to maximize SNR. The results demonstrate a trade-off between the number of allocated subcarriers and the RIS elements required to meet performance targets while also highlighting the influence of RIS quantization levels and the transmitter MIMO array on link quality. Additionally, the proposed PSO algorithm efficiently reduces the number of RIS elements required to establish an RIS-assisted link, resulting in a more energy-efficient setup while preserving the desired signal-to-noise ratio. Finally, future studies may consider the effects of RIS configuration time on channel coherence time under various user mobility speeds, providing a more practical insight into real-world deployment scenarios.
\section*{Acknowledgment}
\textbf{The authors gratefully acknowledge the generous support and donation provided by Nokia, which enabled this collaboration and supported the research presented in this work.}
\balance
\bibliographystyle{IEEEtran}
\bibliography{PHO}

\vspace{-20 mm}

\vskip -2\baselineskip plus -1fil

\end{document}